\pdfoutput=1

\documentclass[11pt]{article}

\usepackage[preprint]{acl}

\usepackage{times}
\usepackage{latexsym}
\usepackage{hyperref}

\usepackage{booktabs}
\usepackage{multirow}
\usepackage{colortbl}
\usepackage{arydshln}

\usepackage{amsmath}
\usepackage{amssymb}
\usepackage{bbm}
\usepackage{bm}
\DeclareMathOperator*{\argmax}{arg\,max}

\usepackage{adjustbox}
\usepackage{subcaption}

\usepackage{xcolor}
\definecolor{backgray}{rgb}{0.9,0.9,0.9}
\definecolor{textgray}{rgb}{0.6,0.6,0.6}
\definecolor{mygreen}{HTML}{CAE5CD}
\definecolor{myred}{HTML}{ffcccc}

\usepackage{tcolorbox}

\newcommand{\xxicl}{X$^\text{2}$-ICL}

\usepackage[T1]{fontenc}

\usepackage[utf8]{inputenc}

\usepackage{microtype}

\usepackage{inconsolata}

\usepackage{graphicx}

\usepackage{dsfont, amsthm, amsfonts, amsmath, bm, bbm}
\newcommand{\Y}{\mathcal{Y}}
\newcommand{\X}{\mathcal{X}}
\newcommand{\LL}{\mathcal{L}}
\newcommand{\R}{\mathcal{R}}
\newcommand{\D}{\mathcal{D}}
\newcommand{\one}{\mathbbm{1}}
\newcommand{\Dmeta}{\mathcal{S}}

\usepackage{algorithm}
\usepackage{algorithmic}
\usepackage{array}
\usepackage{booktabs}
\newcommand{\INPUT}{\STATE \textbf{Input:}}
\newcommand{\OUTPUT}{\STATE \textbf{Output:}}

\title{Exploring Explanations Improves the Robustness of In-Context Learning}

\author{Ukyo Honda \\
  CyberAgent, Tokyo, Japan \\
  \texttt{honda\_ukyo@cyberagent.co.jp} \\\And
  Tatsushi Oka \\
  Keio University, Tokyo, Japan \\
  \texttt{tatsushi.oka@keio.jp} \\}

\begin{document}
\maketitle
\begin{abstract}
In-context learning (ICL) has emerged as a successful paradigm for leveraging large language models (LLMs).
However, it often struggles to generalize beyond the distribution of the provided demonstrations.
A recent advancement in enhancing robustness is ICL with \emph{explanations} (X-ICL), which improves prediction reliability by guiding LLMs to understand and articulate the reasoning behind correct labels.
Building on this approach, we introduce an advanced framework that extends X-ICL by systematically \emph{exploring explanations} for all possible labels (\xxicl{}), thereby enabling more comprehensive and robust decision-making.
Experimental results on multiple natural language understanding datasets validate the effectiveness of \xxicl{}, demonstrating significantly improved robustness to out-of-distribution data compared to the existing ICL approaches.\footnote{The code is publicly available at: \url{https://github.com/CyberAgentAILab/x2-icl}.}
\end{abstract}

\section{Introduction}
\label{sec:intro}
In-context learning \citep[\textbf{ICL};][]{brown2020language,radford2019language} has demonstrated remarkable success with the advancement of large language models (LLMs).
In ICL, LLMs make inferences given a limited number of \emph{demonstrations}, i.e., labeled examples provided as context, without any parameter updates.
This efficiency is a key advantage over the predominant paradigm of fine-tuning, which requires a large amount of task-specific training data and parameter updates.
Not only efficient, but ICL has also been shown to be effective across a variety of tasks, making it a new paradigm for leveraging LLMs \citep{dong-etal-2024-survey}.

However, ICL has been reported to be restricted in its ability to generalize beyond the given demonstrations.
ICL often exhibits degraded performance when evaluated on \emph{out-of-distribution} (OOD) data \citep{tang-etal-2023-large,wang2023adversarial,wang2023on,mueller-etal-2024-context,zhou-etal-2024-explore,siska-etal-2024-examining,yuan-etal-2024-llms}.
Following the previous studies, we specifically refer to OOD data as data drawn from distributions that have been \emph{adversarially shifted} from the distribution of the given demonstrations.
Recent studies have addressed the robustness challenge in ICL \citep{li-etal-2023-robust,sun-etal-2024-exploring,jang-etal-2024-rectifying,zhou-etal-2024-explore,he-etal-2024-using}.

\begin{figure}[t]
\centering
\includegraphics[width=1.0\columnwidth,keepaspectratio]{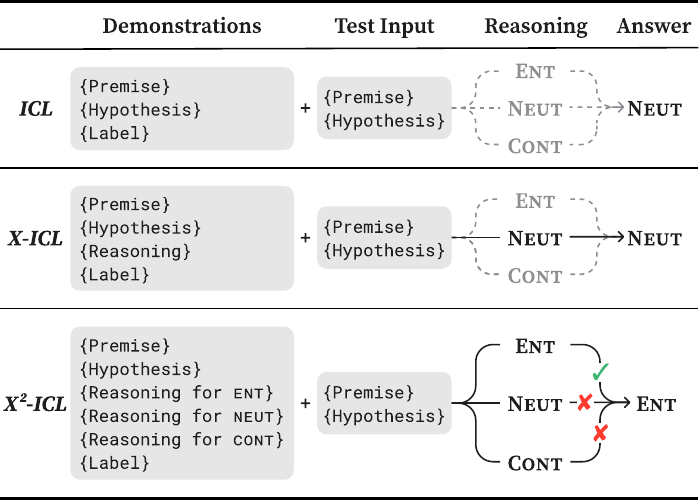}
\caption{
Overview of ICL, X-ICL, and our proposed \xxicl{}.
In this example, the task is natural language inference, where \textsc{Ent}, \textsc{Neut}, and \textsc{Cont} represent entailment, neutral, and contradiction labels, respectively.
Solid lines denote outputs generated by LLMs during inference, while \textcolor{textgray}{dotted gray lines} represent paths not explicitly produced by LLMs.
ICL produces only the final label without providing reasoning, whereas X-ICL generates a single reasoning path before selecting a label.
In contrast, \xxicl{} first generates reasoning for all possible labels and then chooses the label supported by the most valid reasoning.
}
\label{fig:overview}
\end{figure}

A promising advancement in addressing this challenge is ICL with \emph{explanations} \citep[\textbf{X-ICL};][]{he-etal-2024-using}.
In X-ICL, LLMs are provided with labeled examples along with explanations that justify the assigned labels.
The objective is to enable the model to learn human-like \emph{reasoning}\footnote{The explanation that justifies an assigned label represents the reasoning process from the perspective of a model that learns from such explanations. Thus, we use the terms \emph{explanation} and \emph{reasoning} interchangeably, as in \citet{he-etal-2024-using}.} rather than relying on superficial patterns within the provided demonstrations.
This approach is conceptually similar to few-shot chain-of-thought (CoT) reasoning, which provides examples of human-written step-by-step reasoning as context \citep{wei2022chain}.\footnote{While there is a zero-shot CoT approach, which does not provide reasoning examples in the context but only adds an instruction to perform such reasoning \citep{kojima2022large}, X-ICL builds on the few-shot CoT approach.}
However, a technical distinction of X-ICL is that the provided in-context explanations are generated by LLMs, rather than being manually crafted.
This machine-generated nature makes X-ICL scalable, and it has demonstrated strong performance across various OOD evaluation tasks in natural language understanding (NLU).

In this study, we extend X-ICL to enhance robustness by further leveraging its scalability.
In OOD data, the reasoning learned from demonstrations is not always reliable.
As the data distribution is adversarially shifted from that of the demonstrations, models must carefully analyze inputs from multiple aspects to ensure accurate predictions.
Our idea is that \emph{exploring explanations} behind all available labels in demonstrations (\textbf{\xxicl{}}) will encourage the model to more thoroughly consider the alternative reasoning paths.
To achieve this, we augment the demonstrations by providing diverse explanations, each generated by conditioning LLMs on a different possible label.
Figure~\ref{fig:overview} provides an overview of our method.

Our extensive experiments demonstrate the enhanced OOD robustness of \xxicl{} across various datasets in NLU.
Specifically, \xxicl{} outperforms both ICL and X-ICL on six to eight out of eight OOD datasets for each of five different LLMs.
We expect that the effectiveness of our approach will offer important insights into the robustness of ICL, emphasizing the value of systematically and comprehensively exploring various reasoning paths.

\section{Preliminaries}
\label{sec:preliminaries}
To formally introduce existing methods and our proposed method, this section first establishes the notation and problem setup for classification tasks.
Subsequently, we describe how latent variables arise in classification tasks.

\subsection{Notation and Problem Setup}
\label{sec:problem_setup}
Consider a classification problem with random variables
$(x, y)$, where $x$ comprises the predictive features and $y \in \Y$ denotes the class label from a finite set $\Y:=\{1, \dots, L\}$. 
Let $p(x, y)$ and $p(y|x)$ denote the joint and conditional probability distributions, respectively.

For this problem, we define a decision rule 
$\delta: \X \to \Y$ in the functional space $\Delta$
and evaluate it using the binary (0-1) loss function $\LL:\Y \times \Delta \to \{0,1\}$, defined as 
\begin{align}
\LL \big (y, \delta(x) \big)
:= \one \big \{y \not = \delta(x) \big \},    
\end{align}
where $\one\{\cdot\}$ represents the indicator function, taking 1 when the condition specified in the braces is satisfied and 0 otherwise.

Our goal is to find a decision rule $\delta \in \Delta$ that minimizes the the expected loss function, which is written as the misclassification probability:
\begin{align}
\mathbb{E}
\big [
\LL\big(y, \delta(x) \big)
\big ]
=
\Pr\{y \neq \delta(x)   \}. 
\end{align}
Furthermore, the misclassification probability can be expressed through the conditional distribution:
\begin{align}
    \Pr  \{y \neq \delta(x) \}
    =
    \int_{\X} 
    \Pr  \{ y {\neq} \delta(x) |x \}
    p(x) d x. 
\end{align}
This formulation shows that minimizing the overall misclassification error is equivalent to minimizing 
$\Pr  \{ y {\neq} \delta(x) | x \}$ for each $x \in \mathcal{X}$.
Since
$\Pr  \{ y {\neq} \delta(x) | x \} = 1 - 
\Pr  \{ y  {=} \delta(x) | x \}$,
the optimal decision rule $\delta^{\ast} \in \Delta$ takes the form:
\begin{align}
    \delta^{\ast}(x):= \argmax_{y \in \Y}
    p(y|x). 
\end{align}
This decision rule, known as the Bayes optimal classifier, predicts the label $y$ from features $x$ by selecting the one with the highest conditional probability among all labels.

\subsection{Latent-Variable Modeling}
\label{sec:latent}
The classification framework can be extended to incorporate latent variables that model the underlying structure of the data.
Let $r$ be a possibly high-dimensional latent variable that resides in the space $\R$. The likelihood of observed data is represented as:
\begin{align}
p(y|x) = \int_{\R} p(y|r, x)  p(r|x) dr,
\end{align}
where $p(y|r, x)$ describes the enriched conditional probability incorporating the latent information, and 
$p(r|x)$ characterizes the latent probability conditional on observed features. 

This latent variable $r$ encodes additional structure about the relationship between features and labels, such as underlying reasons or mechanisms.
This approach builds on fundamental work in classification analysis \cite{thurstone1927, luce1959, marschak1960, mcfadden1973, dempster1977}, and has since become a cornerstone of modern classification methods.
Importantly, latent variables exist for all possible labels, with the observed label emerging through their interaction. Formally, we define
\begin{align}
    \label{eq:latent}
\bm{r} & := (r_{1}, \dots, r_{L}),
\end{align}
where $r_{\ell}$ denotes the latent element associated with label $\ell \in \Y$.

\section{Methods}
\label{sec:method}
In this section, we first review ICL and X-ICL under the framework introduced in Section~\ref{sec:preliminaries}, and then we introduce \xxicl{}, our proposed method.
Figure~\ref{fig:overview} presents an overview and highlights the differences among these methods.
Hereafter, we refer to a set of \emph{demonstrations} as a training sample $\D_{n}:=\{(x_i, y_i)\}_{i=1}^n$ of $n$ observations independently drawn from
the joint distribution $p(x,y)$.
Our task is to predict the label $y'$ for a new feature variable $x'$ in the test set.

\subsection{ICL}
\label{sec:icl}
ICL emerges as an important paradigm where LLMs learn to perform a task by observing demonstrations within its context window, without updating their parameters \citep{brown2020language,radford2019language}.
The demonstrations $\D_{n}$ are concatenated into a prompt sequence that LLMs process, along with a new query point $x'$, to predict the corresponding label $y'$.

Formally, we can explain the ICL decision rule 
in the likelihood framework as described in Section \ref{sec:problem_setup}. 
LLMs identify and follow patterns in the demonstrations, which corresponds to obtaining a model trained for the task represented by the demonstrations.
We denote the model implicitly learned through ICL as $\hat{p}(y | x)$, which is defined as follows:
\begin{align}
    \hat{p}(y|x)
    := p(y|x,\D_{n}).
\end{align}
Here, we omit the parameters of LLMs for brevity.
The model predicts the most likely label for a new input $x'$ through
\begin{align}
    \delta^{\mathrm{ICL}} (x')
    := \argmax_{y' \in \Y} 
    \hat{p}(y'| x') . 
\end{align}
Algorithm~\ref{alg:icl} in Appendix~\ref{sec:appendix:algo} summarizes the above procedure.

The decision rule $\delta^{\mathrm{ICL}}$ can be regarded as a map from the query input to a label prediction, conditioned on the provided set of demonstrations.
The key distinction is that $\delta^{\mathrm{ICL}}$ learns through pattern recognition within the demonstrations rather than parameter updates.

\subsection{X-ICL: ICL with Explanations}
\label{sec:xicl}
X-ICL incorporates explanations into the demonstrations, thereby guiding LLMs to learn the reasoning process that connects features to labels \citep{he-etal-2024-using}.
This approach can be interpreted as a form of latent-variable modeling, where the explanation serves as the latent variable $r$ that capture the underlying structure governing the relationship between features and labels.

To augment the demonstrations with such explanations, X-ICL leverages LLMs to generate reasoning for each example through few-shot ICL.
For this purpose, X-ICL additionally requires \emph{meta-prompt}, a set of examples annotated with explanations: $\Dmeta_m := \{(x_j, r_j, y_j)\}_{j=1}^m$.
Importantly, the meta-prompt can be quite small; in the tasks examined, a single example per label was sufficient \citep{he-etal-2024-using}.
Therefore, the manual effort required to construct the meta-prompt $\Dmeta_m$ is significantly lower than that needed to manually create explanations for all examples in the demonstrations $\D_n$.

During preprocessing, for each data point in $\D_n$, X-ICL uses $\Dmeta_m$ to collect the reasoning $r_y$ that justifies the assignment of label $y$ to $x$.
Here, $r_y$ can be considered as the $y$-th element of $\bm{r}$ in Eq.~(\ref{eq:latent}).
Formally, X-ICL approximates the conditional probability distribution of explanations as follows:
\begin{align}
     \tilde{p}(r|y, x) := p(r|y, x, \Dmeta_m),    
\end{align}
and then samples reasoning according to:
\begin{align}
     r_{y} \sim \tilde{p}(r|y, x). 
\end{align}
As a result of this augmentation process, the demonstration set in X-ICL becomes $\{(x_i, r_{y_i}, y_i)\}_{i=1}^n$.

In probabilistic terms, we express the joint distribution
of label and its specific latent variable
$p(y, r_{y}|x)$ as     
\begin{align}
    p(y, r_{y}|x) = p(y|r_{y}, x) p(r_{y}|x). 
\end{align}
During ICL with the augmented demonstrations, LLMs learn these two components: the latent variable generation process $\hat{p}(r_{y}|x)$ and the label prediction mechanism $\hat{p}(y|r_{y}, x)$.
At inference time, given a test feature $x' \in \X$, X-ICL first samples a reasoning $r_{y}'$ from $\hat{p}(r_{y}|x)$ and then determine the classification rule:
\begin{align}
    \delta^{\mathrm{X\text{-}ICL}} (x')
    := \argmax_{y' \in \Y} 
    \hat{p}(y'|r_{y}', x') . 
\end{align}
Algorithm~\ref{alg:x-icl} in Appendix~\ref{sec:appendix:algo} summarizes the above procedure.

From the perspective of latent variable modeling, the problem of X-ICL is that it imposes restrictive constraints on the latent space, binding it tightly to realized values in the demonstrations.
This constraint naturally leads to strong in-sample performance but significantly limits generalizability by creating a shallow latent structure that only spans a confined region of the possible latent space.

\subsection{\xxicl{}: ICL with Exploration of Explanations}
\label{sec:xxicl}
To mitigate the constraints imposed in X-ICL, we propose \xxicl{}, a method designed to effectively explore the latent reasoning space.
Unlike conventional approaches that focus solely on the observed label, \xxicl{} examines the underlying rationale that could justify each possible label for the given input features.
By doing so, our approach preserves the dimensionality of the latent reasoning space, allowing for a more comprehensive exploration of potential reasoning paths.

Intuitively, this approach maintains a broader consideration of plausible reasoning paths.
This is particularly crucial in OOD scenarios, where the reasoning learned from demonstrations may not always be reliable, as it is constrained by the observed input--label pairs.
In such cases, alternative reasoning processes, such as those based on different aspects of the input features, may still be valid.
Our method aims to preserve these diverse reasoning paths, thereby enhancing robustness to the distribution shifts.

We employ the same meta-prompt to augment explanations within demonstrations.
However, we extend this approach by augmenting not only the explanations for the observed labels but also those for unrealized labels.
This provides a more comprehensive understanding of the reasoning space.
We draw $r_{\ell}$ separately for each $\ell \in \Y$ as follows:
\begin{align}
     r_{\ell} \sim \tilde{p}(r_{\ell}|y=\ell, x),    
\end{align}
In this step, instead of using only the observed label $y$, we consider what would be reasoning $r$ if the label is $\ell$. 
Considering all potential labels, we can uncover the collection of reasoning $r$ in Eq.~(\ref{eq:latent}). 

By applying the above algorithm to the demonstrations, we obtain an augmented data 
$\{(x_{i}, \bm{r}_{i}, y_{i})\}_{i=1}^{n}$
with $\bm{r}_{i} :=(r_{i,1}, \dots, r_{i,L})$.
Note that $\bm{r}_i$ represents a set of reasoning paths, with each reasoning path $r_{i, \ell}$ corresponding to a specific label $\ell$.
The joint distribution of label and latent variable $p(y, \bm{r}|x)$ is expressed as
\begin{align}
    p(y, \bm{r}|x) = p(y|\bm{r}, x) p(\bm{r}|x). 
\end{align}
Based on the augmented demonstrations, LLMs learn 
the label prediction mechanism $\hat{p}(y|\bm{r}, x)$ 
as well as the joint distribution of latent variables $\hat{p}(\bm{r}|x)$.
For inference, our model takes a test input $x' \in \X$ and then draws a collection of latent variables $\bm{r}'$ from $\hat{p}(\bm{r}|x')$.
Then, we obtain the prediction probability of each label by $\hat{p}(y'|\bm{r}', x')$ for every $y' \in \Y$.
Finally, the classification rule is given by
\begin{align}
    \delta^{\mathrm{X^2\text{-}ICL}}(x')
    := \argmax_{y' \in \Y} 
    \hat{p}(y'|\bm{r}', x') . 
\end{align} 
Algorithm \ref{alg:x2icl} describes this estimation and inference procedure.

\begin{algorithm}[tb]
\caption{\xxicl{}}
\label{alg:x2icl}
\begin{algorithmic}[1]
\INPUT \ $\D_{n}$ and $\Dmeta_{m}$
\FOR{$i = 1, \dots, n $}
\FOR{each label $\ell = 1, \dots, L$}
    \STATE 
    Generate a latent variable
    $r_{i,\ell}$
    for $(x_{i}, \ell)$: \\
    \hspace{1.5cm} $r_{i, \ell}\sim \tilde{p}(r_{\ell}| y_{i}=\ell, x_{i})$
\ENDFOR
    \STATE 
    Obtain a collection of latent variables: \\ 
    \hspace{1.5cm} $\bm{r}_{i}=(r_{i,1}, \dots, r_{i,L})$
\ENDFOR
\STATE \textbf{Estimation:}
\STATE \quad Augumented data: $\{(x_i,\bm{r}_i,y_i)\}_{i=1}^n$
\STATE \quad Model: $p(y, \bm{r}|x)$
\STATE \quad Estimator: 
$\hat{p}(y, \bm{r}|x) = \hat{p}(y|\bm{r}, x) \hat{p}(\bm{r}|x)$
\STATE \textbf{Inference:} 
\STATE \quad Test input $x'$
\STATE \quad Draw $\bm{r}':=(r_{1}', \dots, r_{L}') \sim \hat{p}(\bm{r}|x')$
\STATE \quad Compute 
$\{ \hat{p}(y'| \bm{r}', x'):  y' \in \Y \}$
\OUTPUT \ Classification 
\STATE \quad 
$y^{\ast} = \argmax_{y' \in \mathcal{Y}} \hat{p}(y'| \bm{r}', x')$
\end{algorithmic}
\end{algorithm}

The three methods above represent distinct theoretical principles in their treatment of latent structures.
ICL operates primarily through pattern recognition within observed samples, while X-ICL extends this framework by incorporating reasoning mechanisms for observed labels.
Our approach diverges from these frameworks by adopting a more comprehensive treatment: it systematically uncovers latent reasoning structures for all possible labels, irrespective of whether they are explicitly realized in the observed data.

\begin{table*}[t]
\centering
\begin{adjustbox}{max width=1.0\textwidth}
\begin{tabular}{lcccccccccc}
\toprule
 & \multicolumn{8}{c}{\emph{Natural Language Inference}} & \multicolumn{2}{c}{\emph{Paraphrase}} \\
\cmidrule(lr){2-9}
\cmidrule(lr){10-11}
 & \textcolor{textgray}{\textbf{SNLI}} & \textbf{HANS} & \textbf{NAN} & \textbf{PISP} & \textbf{ST} & \textbf{ANLI$_{\textit{R1}}$} & \textbf{ANLI$_{\textit{R2}}$} & \textbf{ANLI$_{\textit{R3}}$} & \textcolor{textgray}{\textbf{QQP}} & \textbf{PAWS} \\
\midrule
\textbf{GPT-4o} &  &  &  &  &  &  &  &  &  &  \\
ICL & $\textbf{90.95}_{\pm 1.20}$ & $88.05_{\pm 2.52}$ & $75.97_{\pm 0.84}$ & $77.90_{\pm 2.09}$ & $78.25_{\pm 3.64}$ & $70.67_{\pm 2.60}$ & $61.05_{\pm 2.23}$ & $61.58_{\pm 2.32}$ & $\textbf{83.65}_{\pm 2.45}$ & $65.15_{\pm 1.10}$ \\
X-ICL & $90.00_{\pm 0.85}$ & $86.35_{\pm 1.11}$ & $78.29_{\pm 1.23}$ & $81.40_{\pm 1.70}$ & $81.50_{\pm 2.75}$ & $75.58_{\pm 0.92}$ & $63.87_{\pm 0.73}$ & $65.07_{\pm 2.23}$ & $82.75_{\pm 3.19}$ & $63.80_{\pm 3.42}$ \\
\rowcolor{backgray}
\xxicl{} & $90.25_{\pm 0.50}$ & $\textbf{88.85}_{\pm 1.04}$ & $\textbf{78.78}_{\pm 0.37}$ & $\textbf{83.76}_{\pm 0.60}$ & $\textbf{82.35}_{\pm 2.24}$ & $\textbf{77.40}_{\pm 0.48}$ &  $\textbf{67.61}_{\pm 1.89}$ & $\textbf{67.70}_{\pm 1.63}$ & $78.85_{\pm 3.53}$ & $\textbf{70.85}_{\pm 3.88}$ \\
\midrule
\textbf{Gemini-1.5-Pro} &  &  &  &  &  &  &  &  &  &  \\
ICL & $\textbf{89.80}_{\pm 1.07}$ & $86.85_{\pm 1.37}$ & $75.39_{\pm 1.47}$ & $\textbf{80.66}_{\pm 0.97}$ & $84.30_{\pm 0.96}$ & $74.04_{\pm 1.84}$ & $64.48_{\pm 2.30}$ & $64.08_{\pm 2.20}$ & $\textbf{83.00}_{\pm 2.69}$ & $65.20_{\pm 2.87}$ \\
X-ICL & $85.70_{\pm 1.75}$ & $87.80_{\pm 1.55}$ & $\textbf{77.23}_{\pm 1.28}$ & $78.23_{\pm 1.38}$ & $83.35_{\pm 1.96}$ & $73.56_{\pm 1.52}$ & $64.33_{\pm 0.96}$ & $65.59_{\pm 0.95}$ & $82.25_{\pm 2.65}$ & $64.60_{\pm 2.81}$ \\
\rowcolor{backgray}
\xxicl{} & $82.70_{\pm 1.96}$ & $\textbf{88.40}_{\pm 0.94}$ & $76.55_{\pm 1.03}$ & $80.32_{\pm 2.05}$ & $\textbf{85.60}_{\pm 1.36}$ & $\textbf{75.48}_{\pm 1.01}$ & $\textbf{66.77}_{\pm 2.02}$ & $\textbf{67.11}_{\pm 1.31}$ & $80.65_{\pm 2.68}$ & $\textbf{69.80}_{\pm 4.46}$ \\
\midrule
\textbf{Gemini-2.0-Flash} &  &  &  &  &  &  &  &  &  &  \\
ICL & $\textbf{87.15}_{\pm 1.43}$ & $73.05_{\pm 2.37}$ & $74.32_{\pm 0.97}$ & $72.91_{\pm 3.87}$ & $80.45_{\pm 1.89}$ & $69.23_{\pm 2.24}$ & $61.74_{\pm 1.23}$ & $59.54_{\pm 1.93}$ & $81.05_{\pm 3.24}$ & $58.10_{\pm 3.67}$ \\
X-ICL & $84.65_{\pm 1.28}$ & $78.25_{\pm 3.09}$ & $75.97_{\pm 0.32}$ & $75.27_{\pm 1.86}$ & $80.75_{\pm 0.64}$ & $\textbf{75.29}_{\pm 2.58}$ & $63.72_{\pm 1.11}$ & $65.79_{\pm 1.34}$ & $\textbf{82.85}_{\pm 2.37}$ & $56.25_{\pm 7.46}$ \\
\rowcolor{backgray}
\xxicl{} & $84.00_{\pm 2.55}$ & $\textbf{85.40}_{\pm 2.29}$ & $\textbf{76.07}_{\pm 1.07}$ & $\textbf{83.69}_{\pm 1.04}$ & $\textbf{82.50}_{\pm 1.06}$ & $74.90_{\pm 1.31}$ & $\textbf{65.93}_{\pm 1.97}$ & $\textbf{67.37}_{\pm 1.38}$ & $81.55_{\pm 2.24}$ & $\textbf{66.40}_{\pm 4.17}$ \\
\midrule
\midrule
\textbf{Phi-4-14B} &  &  &  &  &  &  &  &  &  &  \\
ICL & $85.45_{\pm 2.82}$ & $80.95_{\pm 2.39}$ & $74.22_{\pm 0.50}$ & $71.63_{\pm 2.22}$ & $74.30_{\pm 1.78}$ & $70.29_{\pm 1.61}$ & $60.29_{\pm 0.29}$ & $61.25_{\pm 1.12}$ & $\textbf{81.10}_{\pm 2.11}$ & $69.45_{\pm 4.98}$ \\
X-ICL & $\textbf{90.00}_{\pm 1.99}$ & $84.05_{\pm 1.90}$ & $\textbf{78.49}_{\pm 1.21}$ & $76.08_{\pm 1.89}$ & $78.15_{\pm 1.43}$ & $73.65_{\pm 2.44}$ & $60.29_{\pm 1.23}$ & $62.37_{\pm 1.89}$ & $80.60_{\pm 3.06}$ & $68.20_{\pm 5.82}$ \\
\rowcolor{backgray}
\xxicl{} & $80.45_{\pm 2.24}$ & $\textbf{84.45}_{\pm 1.06}$ & $77.71_{\pm 1.43}$ & $\textbf{78.37}_{\pm 1.23}$ & $\textbf{80.15}_{\pm 1.72}$ & $\textbf{75.38}_{\pm 1.40}$ & $\textbf{63.34}_{\pm 1.46}$ & $\textbf{63.42}_{\pm 1.23}$ & $70.75_{\pm 4.37}$ & $\textbf{75.90}_{\pm 3.61}$ \\
\midrule
\textbf{DeepSeek-R1-8B} &  &  &  &  &  &  &  &  &  &  \\
ICL & $59.55_{\pm 1.53}$ & $74.65_{\pm 3.64}$ & $64.73_{\pm 3.20}$ & $62.74_{\pm 3.59}$ & $66.00_{\pm 1.98}$ & $55.87_{\pm 1.70}$ & $49.09_{\pm 2.89}$ & $49.87_{\pm 1.85}$ & $74.25_{\pm 2.18}$ & $\textbf{48.55}_{\pm 2.30}$ \\
X-ICL & $63.20_{\pm 1.18}$ & $78.45_{\pm 1.98}$ & $69.38_{\pm 1.64}$ & $67.25_{\pm 1.35}$ & $67.40_{\pm 1.07}$ & $60.87_{\pm 2.58}$ & $50.53_{\pm 3.98}$ & $\textbf{50.59}_{\pm 1.64}$ & $\textbf{76.60}_{\pm 1.41}$ & $46.60_{\pm 3.72}$ \\
\rowcolor{backgray}
\xxicl{} & $\textbf{66.10}_{\pm 2.08}$ & $\textbf{81.65}_{\pm 2.77}$ & $\textbf{70.45}_{\pm 1.11}$ & $\textbf{68.73}_{\pm 1.96}$ & $\textbf{68.10}_{\pm 1.55}$ & $\textbf{63.56}_{\pm 2.95}$ & $\textbf{52.13}_{\pm 2.30}$ & $49.67_{\pm 1.46}$ & $74.15_{\pm 0.72}$ & $41.95_{\pm 2.80}$ \\
\bottomrule
\end{tabular}
\end{adjustbox}
\caption{
Main results. The scores are the mean and standard deviation of accuracy computed across four different sets of demonstrations.
The highest mean score for each model is highlighted in bold.
\textcolor{textgray}{\textbf{SNLI}} and \textcolor{textgray}{\textbf{QQP}} share the same data distribution as the demonstrations, whereas the other datasets are OOD.
}
\label{tab:main_results}
\end{table*}

\section{Experiments}
\label{sec:experiments}
In this section, we evaluate the OOD robustness of \xxicl{}, primarily through comparison with ICL and X-ICL across diverse NLU datasets.

\subsection{Experimental Setup}
\label{sec:setup}

\paragraph{Tasks.}
\label{sec:tasks}
We employed two NLU tasks: natural language inference (NLI) and paraphrase identification.
NLI is the task of classifying the semantic relationship between a hypothesis and a premise into one of three categories: \emph{entailment}, \emph{neutral}, or \emph{contradiction}.
Paraphrase identification is the task of determining whether a given pair of sentences has the same meaning, classifying them as either \emph{yes} or \emph{no}.
Examples for each label are shown in Figures~\ref{fig:xprompt_example_nli} and \ref{fig:xprompt_example_qqp} in Appendix~\ref{sec:appendix:prompts}.

\paragraph{Datasets.}
\label{sec:datasets}
We used OOD evaluation datasets for NLI and paraphrase identification.
To ensure the correctness of the labels, we selected datasets that were either manually curated and thoroughly verified or generated based on well-defined rules.
In particular, for NLI, we employed HANS \citep{mccoy-etal-2019-right}, NAN \citep{truong-etal-2022-another}, PISP \citep{liu-etal-2020-hyponli,liu-etal-2020-empirical}, ST \citep{naik-etal-2018-stress}, and ANLI$_\textit{R1--R3}$ \citep{nie-etal-2020-adversarial}, and for paraphrase identification, we used PAWS \citep{zhang-etal-2019-paws}.
We used SNLI training data \citep{bowman-etal-2015-large} as demonstrations for evaluation on the NLI test sets, ANLI$_\textit{R1}$ training data for the ANLI test sets, and QQP training data \citep{qqp2017} for PAWS, following X-ICL \citep{he-etal-2024-using}.\footnote{
Additional reason for using ANLI trainig data is that ANLI premises are generally much longer than those in other NLI datasets.
This clear difference could interfere with evaluating the robustness that ANLI is intended to measure, namely, robustness to subtle changes that challenge models but are less noticeable to humans.
To accurately assess the distribution shift targeted by ANLI, we avoid using demonstrations from other NLI datasets.
Note that the test sets of ANLI are considered OOD even with respect to ANLI training data, due to differences in the rounds of dataset creation and in the annotators, which are known sources of distribution shifts \citep{nie-etal-2020-adversarial,gururangan-etal-2018-annotation,geva-etal-2019-modeling}.
}
Although ICL does not involve updating the parameters of LLMs, it enables LLMs to learn patterns from demonstrations provided in the context.
The evaluation data above are designed to reflect adversarial distribution shifts relative to the training data, making them suitable benchmarks for evaluating the OOD robustness of ICL.
All the datasets are in English and are publicly available for model evaluation.
See Appendix \ref{sec:appendix:datasets} for more details of the datasets.

\paragraph{Evaluation.}
\label{sec:evaluation}
We followed the evaluation protocol of X-ICL \citep{he-etal-2024-using}.
The number of in-context demonstrations was set to eight, which is commonly referred to as the 8-shot ICL setting.
We used four different random seeds to select both demonstrations and test samples, sampling up to 500 test examples from each test set \citep{jang-etal-2024-rectifying}.
The reported scores correspond to the accuracy averaged over the four random seeds.

\paragraph{Models.}
\label{sec:model}
We used three cutting-edge LLMs: GPT-4o (\texttt{gpt-4o-2024-08-06}), Gemini-1.5-Pro (\texttt{gemini-1.5-pro-002}), and Gemini-2.0-Flash (\texttt{gemini-2.0-flash-001}).
To assess the broader applicability of our approach, we additionally employed more computationally efficient open-source models: Phi-4-14B\footnote{\url{https://huggingface.co/unsloth/phi-4}} \citep{abdin2024phi} and DeepSeek-R1-8B\footnote{\url{https://huggingface.co/deepseek-ai/DeepSeek-R1-Distill-Llama-8B}} \citep{guo2025deepseek}.
Appendix \ref{sec:appendix:models} provides details on decoding configurations.

\paragraph{Explanations.}
\label{sec:explanations}
To generate explanations for incorporation into demonstrations, we employed the few-shot meta-prompts introduced by X-ICL \citep{he-etal-2024-using}.
The exact meta-prompts used in our experiments are provided in Figures~\ref{fig:xprompt_example_nli} and \ref{fig:xprompt_example_qqp} in Appendix~\ref{sec:appendix:prompts}.
These meta-prompts include only a single explanation per label, which minimizes the manual effort required to create explanation examples in advance.
In all experimental settings, we used GPT-4o to generate the explanations incorporated into the demonstrations.
Examples of \xxicl{}'s prompts are presented in Figures~\ref{fig:prompt_xxicl} and \ref{fig:prompt_xxicl_qqp} in Appendix~\ref{sec:appendix:prompts}.

\subsection{Main Results}
\label{sec:main_results}
Table~\ref{tab:main_results} shows that X-ICL outperforms ICL on OOD datasets, and \xxicl{} further enhances OOD performance beyond X-ICL overall.
The improvement is more pronounced when employing the higher-performing LLMs.
This suggests that \xxicl{} relies on strong reasoning capabilities to be effective.
Unlike X-ICL, which requires only a single reasoning path, \xxicl{} explores multiple reasoning paths.
Consequently, a higher reasoning capability is essential for its effectiveness.

In contrast, X-ICL and \xxicl{} exhibit degraded performance on SNLI and QQP, which share the same underlying distribution as the demonstrations.
Due to the identical underlying distribution, simply following the patterns present in the demonstrations results in strong performance.
However, the augmented explanations are drawn from LLMs and may not follow the exact distribution of the original input--label pairs.
This deviation could lead to the degradation of the explanation-based methods in performance on the in-distribution data.
This trade-off between in-distribution and OOD performance remains a challenge to be addressed in future work.
However, given the consistent effectiveness of our approach across a diverse range of OOD datasets, we believe that our method makes a significant contribution to enhancing the robustness of ICL.

\subsection{Comparison with Retrieval-Based ICL}
\label{sec:retrieve}

\begin{table*}[t]
\centering
\begin{adjustbox}{max width=1.0\textwidth}
\begin{tabular}{lcccccccccc}
\toprule
 & \multicolumn{8}{c}{\emph{Natural Language Inference}} & \multicolumn{2}{c}{\emph{Paraphrase}} \\
\cmidrule(lr){2-9}
\cmidrule(lr){10-11}
 & \textcolor{textgray}{\textbf{SNLI}} & \textbf{HANS} & \textbf{NAN} & \textbf{PISP} & \textbf{ST} & \textbf{ANLI$_{\textit{R1}}$} & \textbf{ANLI$_{\textit{R2}}$} & \textbf{ANLI$_{\textit{R3}}$} & \textcolor{textgray}{\textbf{QQP}} & \textbf{PAWS} \\
\midrule
\textbf{GPT-4o} &  &  &  &  &  &  &  &  &  &  \\
Set-BSR & $\textbf{90.60}_{\pm 0.83}$ & $85.40_{\pm 2.05}$ & $77.62_{\pm 0.66}$ & $79.99_{\pm 0.92}$ & $77.95_{\pm 1.08}$ & $74.42_{\pm 0.80}$ & $58.69_{\pm 0.18}$ & $61.18_{\pm 0.46}$ & $\textbf{85.20}_{\pm 1.40}$ & $\textbf{72.25}_{\pm 2.11}$ \\
\cdashline{1-11}
X-ICL & $90.00_{\pm 0.85}$ & $86.35_{\pm 1.11}$ & $78.29_{\pm 1.23}$ & $81.40_{\pm 1.70}$ & $81.50_{\pm 2.75}$ & $75.58_{\pm 0.92}$ & $63.87_{\pm 0.73}$ & $65.07_{\pm 2.23}$ & $82.75_{\pm 3.19}$ & $63.80_{\pm 3.42}$ \\
\rowcolor{backgray}
\xxicl{} & $90.25_{\pm 0.50}$ & $\textbf{88.85}_{\pm 1.04}$ & $\textbf{78.78}_{\pm 0.37}$ & $\textbf{83.76}_{\pm 0.60}$ & $\textbf{82.35}_{\pm 2.24}$ & $\textbf{77.40}_{\pm 0.48}$ &  $\textbf{67.61}_{\pm 1.89}$ & $\textbf{67.70}_{\pm 1.63}$ & $78.85_{\pm 3.53}$ & $70.85_{\pm 3.88}$ \\
\midrule
\midrule
\textbf{Phi-4-14B} &  &  &  &  &  &  &  &  &  &  \\
Set-BSR & $87.65_{\pm 1.64}$ & $83.35_{\pm 2.22}$ & $76.94_{\pm 1.50}$ & $71.50_{\pm 1.53}$ & $76.30_{\pm 1.28}$ & $\textbf{76.54}_{\pm 0.00}$ & $56.40_{\pm 0.00}$ & $58.42_{\pm 0.00}$ & $78.00_{\pm 2.23}$ & $75.07_{\pm 1.76}$ \\
\cdashline{1-11}
X-ICL & $\textbf{90.00}_{\pm 1.99}$ & $84.05_{\pm 1.90}$ & $\textbf{78.49}_{\pm 1.21}$ & $76.08_{\pm 1.89}$ & $78.15_{\pm 1.43}$ & $73.65_{\pm 2.44}$ & $60.29_{\pm 1.23}$ & $62.37_{\pm 1.89}$ & $\textbf{80.60}_{\pm 3.06}$ & $68.20_{\pm 5.82}$ \\
\rowcolor{backgray}
\xxicl{} & $80.45_{\pm 2.24}$ & $\textbf{84.45}_{\pm 1.06}$ & $77.71_{\pm 1.43}$ & $\textbf{78.37}_{\pm 1.23}$ & $\textbf{80.15}_{\pm 1.72}$ & $75.38_{\pm 1.40}$ & $\textbf{63.34}_{\pm 1.46}$ & $\textbf{63.42}_{\pm 1.23}$ & $70.75_{\pm 4.37}$ & $\textbf{75.90}_{\pm 3.61}$ \\
\bottomrule
\end{tabular}
\end{adjustbox}
\caption{
\looseness=-1
Comparison between retrieval-based ICL and explanation-based ICL.
The top side of the dotted line corresponds to retrieval-based ICL, and the bottom side to explanation-based ICL.
The notation follows that of Table~\ref{tab:main_results}.
}
\label{tab:retrieval_results}
\end{table*}

While our method explores the reasoning space to achieve robust predictions, one might wonder whether exploring the demonstration space could achieve similar results.
In ICL, selecting demonstrations that are semantically similar to a test input has been shown to perform well \citep{liu-etal-2022-makes}.
Subsequent research has extended the retrieval-based approach to enhance the diversity of selected demonstrations \citep{levy-etal-2023-diverse,pmlr-v202-ye23c,gupta-etal-2023-coverage}.
To assess the effectiveness of exploring the demonstration space in OOD settings, we conducted a comparison with diversity-aware retrieval-based ICL.

We selected Set-BSR \citep{gupta-etal-2023-coverage} for comparison.
Set-BSR uses BERTScore \citep{Zhang2020BERTScore:} to measure semantic similarity and retrieves a diverse set of demonstrations that collectively cover various aspects of similarity to the test inputs.
We employed one model each from the closed-source and open-source groups.
In Table~\ref{tab:retrieval_results}, we observed that Set-BSR improves performance in some cases, but overall, its performance improvements on OOD datasets are notably smaller than those achieved by the explanation-based ICL methods.
This aligns with the original findings in the X-ICL paper \citep{he-etal-2024-using}, which also reported stronger OOD robustness for explanation-based ICL compared to retrieval-based ICL.
These results suggest that diversity in the reasoning space contributes significantly more to OOD robustness than diversity in the demonstration space.\footnote{
These two types of diversity are not necessarily equivalent.
For instance, even if the surface forms or topics of two premise--hypothesis pairs differ greatly, the underlying reasoning can still be identical, e.g., ``the hypothesis contains information that is not explicitly mentioned in the premise, so the label is neutral.''
}

Additionally, we emphasize that explanation-based ICL and retrieval-based ICL are designed for different scenarios.
Retrieval-based ICL typically requires access to large numbers of labeled examples to select effective demonstrations, while explanation-based ICL needs far fewer labeled examples and explanation annotations.\footnote{
For Set-BSR, we followed the setup described by \citet{gupta-etal-2023-coverage} and used up to 44,000 demonstrations.
For X-ICL and \xxicl{}, we used only eight demonstrations, in addition to a single explanation-annotated demonstration for each label, resulting in a total of 8 + 3 demonstrations in the case of NLI.
}
Thus, retrieval-based ICL is suitable when there is plenty of labeled data, whereas explanation-based ICL is effective when only minimal annotation is feasible.

\subsection{Ablation Study}
\label{sec:ablation}
In this ablation study, we investigate the key factors contributing to the improved OOD performance of \xxicl{}.
Specifically, we used GPT-4o to evaluate performance on the ANLI test sets, which are the most challenging among the OOD datasets in our main experiments.
This level of difficulty is well-suited for detailed evaluation of the robustness of high-performing LLMs, as less challenging datasets may not reveal subtle differences.

\paragraph{($\mathrm{\textbf{i}}$) Systematic Exploration of Reasoning.}
Self-consistency \citep[SC;][]{wang2023selfconsistency}, which stochastically samples multiple outputs from models, shares similarities with \xxicl{} in its ability to explore multiple reasoning paths.
However, the key distinction lies in the structured nature of \xxicl{}'s exploration: \xxicl{} systematically considers reasoning paths for all possible labels, whereas SC is constrained by the reasoning paths present between the given inputs and observed labels.
To assess the impact of this systematic exploration, we compare SC with \xxicl{}.
For a fair comparison, we align the number of reasoning paths in SC with the number of labels (three in NLI) and set the sampling temperature to 0.7, following \citet{wang2023selfconsistency}.
As shown in Table~\ref{tab:ablation_reason} ($\mathrm{i}$), \xxicl{} consistently outperforms SC across all the ANLI test sets, indicating that the structured exploration is crucial in improving OOD robustness.

\begin{table}[t]
\centering
\begin{adjustbox}{max width=1.0\columnwidth}
\begin{tabular}{lccc}
\toprule
 & \textbf{ANLI$_{\textit{R1}}$} & \textbf{ANLI$_{\textit{R2}}$} & \textbf{ANLI$_{\textit{R3}}$} \\
\midrule
X-ICL & $75.58_{\pm 0.92}$ & $63.87_{\pm 0.73}$ & $65.07_{\pm 2.23}$ \\
\midrule
\textbf{Ablation} &  &  &  \\
($\mathrm{i}$) X-ICL + SC & $75.96_{\pm 0.80}$ & $65.62_{\pm 1.33}$ & $65.53_{\pm 2.38}$ \\
($\mathrm{ii}$) X-ICL + Instruction & $74.52_{\pm 1.38}$ & $64.33_{\pm 1.55}$ & $65.39_{\pm 3.16}$ \\
($\mathrm{iii}$-a) ICL + CoT & $74.90_{\pm 4.37}$ & $65.55_{\pm 1.67}$ & $65.99_{\pm 0.84}$ \\
($\mathrm{iii}$-b) ICL + Explore CoT & $74.42_{\pm 3.21}$ & $65.55_{\pm 3.07}$ & $65.66_{\pm 2.60}$ \\
\midrule
\rowcolor{backgray}
\xxicl{} & $\textbf{77.40}_{\pm 0.48}$ &  $\textbf{67.61}_{\pm 1.89}$ & $\textbf{67.70}_{\pm 1.63}$ \\
\bottomrule
\end{tabular}
\end{adjustbox}
\caption{
Ablation study examining the contribution of systematic reasoning exploration.
The results for GPT-4o are reported.
The notation follows that of Table~\ref{tab:main_results}.
}
\label{tab:ablation_reason}
\end{table}

\paragraph{($\mathrm{\textbf{ii}}$) Instruction for Exploration.}
\xxicl{} incorporates an instruction designed to guide reasoning exploration:
\begin{quote}
\emph{Instruction: Explore the reasoning behind all the labels. Then, select the label that has the most valid reasoning.}
\end{quote}
See Figures~\ref{fig:prompt_xxicl} and \ref{fig:prompt_xxicl_qqp} in Appendix~\ref{sec:appendix:prompts} for examples.
This instruction serves to clarify the purpose of providing multiple reasoning paths, and additionally, implicitly conveys that the entire label space is covered within these reasoning paths.
A potential concern is that this instruction might offer an unfair advantage over X-ICL, as previous research reported that presenting information about the label space enhanced ICL performance \citep{min-etal-2022-rethinking}.
To isolate the effect of this implicit label-space information from our overall method, we conducted a comparison with X-ICL augmented with a comparable instruction:
\begin{quote}
\emph{Instruction: Explain the reasoning and then select the correct label from entailment, neutral, or contradiction.}
\end{quote}
As shown in Table~\ref{tab:ablation_reason} ($\mathrm{ii}$), adding this instruction to X-ICL does not lead to significant performance improvements.
The results indicate that the performance gains in \xxicl{} arise from the reasoning exploration encouraged by the instruction, rather than from the implicit label-space information.

\paragraph{($\mathrm{\textbf{iii}}$) Explanations within Demonstrations.}
We evaluated the necessity of including explanations within the demonstrations.
This analysis is not only for \xxicl{} but also for X-ICL.
LLMs possess strong reasoning capabilities, allowing them to engage in reasoning processes even without explicit demonstrations of how to do so within the given context \citep{kojima2022large}.
To assess the benefit of X-ICL and \xxicl{} over this zero-shot reasoning, we appended the CoT prompt to the end of the input, without providing any explanations within the demonstrations \citep{kojima2022large}.
Additionally, we included a sentence to ensure adherence to the specified output format.
The full CoT prompt is as follows:
\begin{quote}
\emph{Let's think step by step. Ensure that your response ends with Label: followed by your final answer.}
\end{quote}
The results in Table~\ref{tab:ablation_reason} ($\mathrm{iii}$-a) indicate that the performance of this zero-shot CoT is comparable to that of X-ICL, suggesting that presenting a single reasoning path per demonstration does not provide substantial additional information to the cutting-edge LLM.
However, \xxicl{} still outperforms this baseline, demonstrating that explicitly guiding LLMs to explore multiple reasoning paths in a systematic manner remains beneficial for enhancing the OOD robustness.

Next, we investigated whether reasoning paths can be systematically explored without providing demonstrations of such reasoning.
To this end, we modified the above CoT prompt as follows:
\begin{quote}
\emph{Let's think step by step, exploring the reasons why each label could be correct. Ensure that your response ends with Label: and your final answer.}
\end{quote}
The results in Table~\ref{tab:ablation_reason} ($\mathrm{iii}$-b) show no significant performance difference compared to ($\mathrm{iii}$-a).
Upon inspection, we found that the model often failed to provide label-conditioned reasoning and instead produced generic step-by-step explanations.
This shows that even advanced LLMs like GPT-4o struggle to generate comprehensive reasoning without demonstrations.
Explanations in demonstrations are essential to guide LLMs in performing structured, comprehensive reasoning.

\paragraph{}
The results in ($\mathrm{i}$)--($\mathrm{iii}$) show that the robustness of \xxicl{} arises from the systematic exploration of explanations learned from demonstrations.

\subsection{Qualitative Analysis}
\label{sec:analysis}
We analyzed both the successful and failure cases of \xxicl{} to gain a deeper understanding of its behavior.
Specifically, we examined the outputs generated by GPT-4o on the ANLI$_{\textit{R1}}$ test set.
Due to space limitations, the outputs analyzed in this section are provided in Appendix~\ref{sec:appendix:outputs}.

\paragraph{Successful Cases.}
In summary, \xxicl{} tends to guide the model to carefully examine the details of the input, thereby facilitating reasoning grounded in confidently inferable information.
In contrast, X-ICL often draws conclusions from the absence of explicit information in the input.
For instance, as illustrated in Figure~\ref{fig:output_pos_1}, \xxicl{} deduced that a person whose name is associated with an equation is likely influential, even though this is not explicitly stated in the premise.
Similarly, in Figure~\ref{fig:output_pos_2}, \xxicl{} correctly inferred the end date of an American football player's career by performing a calculation based on details provided in the input, despite the end date not being explicitly mentioned.
In both examples, X-ICL incorrectly assigned the neutral label based on the absence of explicit mention of the above information.
In Addition, \xxicl{} was also effective in preventing the model from making hasty inferences by highlighting the absence of decisive information within the given context, as seen in Figures~\ref{fig:output_pos_3} and \ref{fig:output_pos_4}.

Among the 42 test instances where the answers of X-ICL and \xxicl{} differed, there were 24 cases where X-ICL predicted neutral, but \xxicl{} predicted entailment or contradiction. In contrast, only in 14 instances, X-ICL predicted a label other than neutral, but \xxicl{} predicted neutral.
This suggests that \xxicl{} has a greater tendency to analyze the input more closely and draw on inferable information when making decisions.

\paragraph{Failure Cases.}
While generally beneficial, such in-depth consideration of details does not always lead to correct answers.
For example, as seen in Figure~\ref{fig:output_neg_1}, \xxicl{} incorrectly inferred a driver's race participation history based on his current career.
In some cases, the reasoning appears plausible but does not match the gold label.
In Figure~\ref{fig:output_neg_2}, \xxicl{} concluded that the language of a film was not English because it was awarded in the Academy Award for Best Foreign Language Film category.
While this reasoning seems reasonable, it contradicts the gold label.
Additionally, \xxicl{} sometimes relies on encyclopedic knowledge beyond the given premise, which is not intended for this task, leading to incorrect predictions.
Figure~\ref{fig:output_neg_3} shows an example where \xxicl{} made inappropriate use of encyclopedic knowledge about a film director.

\looseness=-1
In 7 out of 100 test instances, X-ICL made correct predictions while \xxicl{} failed.
Of these, three involved \xxicl{} using encyclopedic knowledge and two involved inferred information.
In contrast, the remaining two were answered based on the absence of information, as shown in Figure~\ref{fig:output_neg_4}.

\paragraph{}
The in-depth reasoning that incorporates more details is a key characteristic of \xxicl{}.
Overall, as the experimental results demonstrate, this reasoning strategy enhances OOD robustness.

\section{Related Work}
\label{sec:related_work}
\paragraph{OOD Robustness in ICL.}
Prior to X-ICL \citep{he-etal-2024-using}, several approaches were explored that did not incorporate explanations.
\citet{sun-etal-2024-exploring} introduced a method using random labels and minimal task descriptions to mitigate spurious correlations between labels and tasks.
In Appendix~\ref{sec:appendix:expt_mixp}, we show that explanation-based approaches substantially outperform this approach.
\citet{jang-etal-2024-rectifying} calibrated model output probabilities to reduce reliance on semantic priors within demonstrations.
However, this calibration requires access to output probabilities, which critically limits its applicability to cutting-edge LLMs, which are often closed-source.
\citet{zhou-etal-2024-explore} proposed balancing the concepts within demonstrations, but this balancing requires prior identification of potentially harmful concepts, which is an impractical requirement.
This requirement has been acknowledged in the literature on OOD robustness as a significant practical limitation \citep{clark-etal-2019-dont,liu2021just,pmlr-v139-creager21a,pmlr-v202-yang23s,honda-etal-2024-eliminate}.
\citet{li-etal-2023-robust} applied a prompt optimization method in OOD settings.
However, their approach presupposes that the target distribution has already been identified and that a batch of unlabeled examples from that same distribution is available.
Moreover, the optimization process involves repeated sampling from LLMs, which can be computationally expensive.

Unlike the above approaches, X-ICL and \xxicl{} achieve strong OOD performance without requiring knowledge of the underlying distribution or optimization, working effectively for both open-source and closed-source models.

\paragraph{ICL with Explanations.}
CoT reasoning has been shown to significantly improve ICL in both few-shot \citep{wei2022chain} and zero-shot settings \cite{kojima2022large}.
Self-consistency \citep[SC;][]{wang2023selfconsistency} extends this approach by sampling multiple reasoning paths and aggregating their outcomes to improve overall performance.
Our \xxicl{} also explores multiple reasoning paths; however, it differs from SC in that it systematically traverses these paths rather than relying on stochastic sampling.
In Section~\ref{sec:ablation}~($\mathrm{i}$), we have demonstrated that this structured exploration leads to improved OOD generalization compared to SC.

\paragraph{OOD Robustness prior to ICL.}
\looseness=-1
OOD robustness has been one of the major challenges in machine learning, with extensive research conducted within the fine-tuning paradigm before the rise of ICL.
Principal existing approaches include reweighting \citep{clark-etal-2019-dont, he-etal-2019-unlearn, karimi-mahabadi-etal-2020-end}, distributionally robust optimization \citep{sagawa2020distributionally}, and invariant risk minimization \citep{arjovsky2019invariant}.
The approach most similar to X-ICL is the one that leverages human-annotated rationales \citep{chen-etal-2022-rationalization,stacey2022supervising,kavumba-etal-2023-prompting,ludan-etal-2023-explanation}.
Analogous to \xxicl{}, \citet{honda-etal-2024-eliminate} modeled various latent features and utilized them for prediction.
Although these fine-tuning methods share conceptual similarities with approaches for improving ICL robustness, their reliance on extensive training data and parameter updates makes them unsuitable for direct use in ICL.

\section{Conclusion}
\label{sec:conclusion}
In this study, we introduced \xxicl{}, a novel framework designed to improve the robustness of ICL on OOD data.
Our method extends X-ICL by incorporating a reasoning mechanism that systematically explores explanations for all possible labels, enabling a more comprehensive exploration of the reasoning space.
Through extensive experiments, we have demonstrated that \xxicl{} exhibits strong robustness compared to both standard ICL and X-ICL, particularly in handling OOD scenarios.
We believe that our findings, which emphasize the importance of explicit and comprehensive exploration of reasoning, represent a significant step toward developing robust ICL methods.

\section*{Limitations}
\label{sec:limitations}
\xxicl{} relies on a fixed label space to systematically explore the reasoning underlying the decision-making process.
Thus, it is not directly applicable to open-ended tasks and may be inefficient when handling tasks with an excessively large label set.

As observed in Section~\ref{sec:main_results}, \xxicl{} exhibits lower performance on in-distribution datasets while demonstrating strong robustness to OOD datasets.
Addressing this performance trade-off remains an open challenge.
For example, investigating the adaptive use of \xxicl{} depending on the inputs to mitigate the trade-off would be an interesting direction for future work.
Overall, we emphasize that this study represents a significant step toward improving OOD robustness in ICL.

One possible reason for the trade-off is the inherent label ambiguity in standard NLI datasets \citep{10.1162/tacl_a_00293,nie-etal-2020-learn,10.1162/tacl_a_00523}.
The practice of single-label assignment in these datasets, without consideration of the ambiguity, may amplify distributional differences between datasets.
Future work includes investigating additional-label or multi-label prediction approaches \citep{10.1162/tacl_a_00523} on ambiguity-aware datasets such as ChaosNLI \citep{nie-etal-2020-learn}, which may help disentangle the impact of label ambiguity from the trade-off.

\xxicl{} requires models with advanced reasoning capabilities to function effectively, as discussed in Section~\ref{sec:main_results}.
This necessity arises from the fact that our approach involves generating and evaluating multiple reasoning paths that are consistent with the given input.
While this reliance on high-level reasoning presents a limitation,  we believe that the rapid advancements in LLMs will progressively alleviate this issue over time.

Due to the comprehensive reasoning during inference, \xxicl{} incurs higher costs than ICL and X-ICL.
Appendix~\ref{sec:appendix:cost} shows the comparison of the costs.
This is another limitation of our approach and reducing the costs will be our future work.
However, we emphasize that the primary focus and contribution of this study lies in proposing the effective ICL method for addressing OOD data, rather than improving computational efficiency.

\section*{Ethical Considerations}
\label{sec:ethical}
LLMs are known to exhibit biases toward certain social groups \citep{pmlr-v202-santurkar23a}, which can inadvertently affect their outputs.
However, we do not anticipate that our proposed method will reinforce such biases.
\xxicl{} facilitates the exploration of diverse reasoning paths, which may contribute to mitigating inherent biases.

\bibliography{acl2025}

\begin{thebibliography}{59}
\providecommand{\natexlab}[1]{#1}

\bibitem[{Abdin et~al.(2024)Abdin, Aneja, Behl, Bubeck, Eldan, Gunasekar, Harrison, Hewett, Javaheripi, Kauffmann et~al.}]{abdin2024phi}
Marah Abdin, Jyoti Aneja, Harkirat Behl, S{\'e}bastien Bubeck, Ronen Eldan, Suriya Gunasekar, Michael Harrison, Russell~J Hewett, Mojan Javaheripi, Piero Kauffmann, and 1 others. 2024.
\newblock \href {https://arxiv.org/abs/2412.08905} {Phi-4 technical report}.
\newblock \emph{arXiv preprint arXiv:2412.08905v1}.

\bibitem[{Arjovsky et~al.(2019)Arjovsky, Bottou, Gulrajani, and Lopez-Paz}]{arjovsky2019invariant}
Martin Arjovsky, L{\'e}on Bottou, Ishaan Gulrajani, and David Lopez-Paz. 2019.
\newblock \href {https://doi.org/10.48550/arXiv.1907.02893} {Invariant risk minimization}.
\newblock \emph{arXiv preprint arXiv:1907.02893v3}.

\bibitem[{Bowman et~al.(2015)Bowman, Angeli, Potts, and Manning}]{bowman-etal-2015-large}
Samuel~R. Bowman, Gabor Angeli, Christopher Potts, and Christopher~D. Manning. 2015.
\newblock \href {https://doi.org/10.18653/v1/D15-1075} {A large annotated corpus for learning natural language inference}.
\newblock In \emph{Proceedings of the 2015 Conference on Empirical Methods in Natural Language Processing}, pages 632--642, Lisbon, Portugal. Association for Computational Linguistics.

\bibitem[{Brown et~al.(2020)Brown, Mann, Ryder, Subbiah, Kaplan, Dhariwal, Neelakantan, Shyam, Sastry, Askell, Agarwal, Herbert-Voss, Krueger, Henighan, Child, Ramesh, Ziegler, Wu, Winter, Hesse, Chen, Sigler, Litwin, Gray, Chess, Clark, Berner, McCandlish, Radford, Sutskever, and Amodei}]{brown2020language}
Tom Brown, Benjamin Mann, Nick Ryder, Melanie Subbiah, Jared~D Kaplan, Prafulla Dhariwal, Arvind Neelakantan, Pranav Shyam, Girish Sastry, Amanda Askell, Sandhini Agarwal, Ariel Herbert-Voss, Gretchen Krueger, Tom Henighan, Rewon Child, Aditya Ramesh, Daniel Ziegler, Jeffrey Wu, Clemens Winter, and 12 others. 2020.
\newblock \href {https://proceedings.neurips.cc/paper_files/paper/2020/file/1457c0d6bfcb4967418bfb8ac142f64a-Paper.pdf} {Language models are few-shot learners}.
\newblock In \emph{Advances in Neural Information Processing Systems}, volume~33, pages 1877--1901. Curran Associates, Inc.

\bibitem[{Chen et~al.(2022)Chen, He, Narasimhan, and Chen}]{chen-etal-2022-rationalization}
Howard Chen, Jacqueline He, Karthik Narasimhan, and Danqi Chen. 2022.
\newblock \href {https://doi.org/10.18653/v1/2022.naacl-main.278} {Can rationalization improve robustness?}
\newblock In \emph{Proceedings of the 2022 Conference of the North American Chapter of the Association for Computational Linguistics: Human Language Technologies}, pages 3792--3805, Seattle, United States. Association for Computational Linguistics.

\bibitem[{Clark et~al.(2019)Clark, Yatskar, and Zettlemoyer}]{clark-etal-2019-dont}
Christopher Clark, Mark Yatskar, and Luke Zettlemoyer. 2019.
\newblock \href {https://doi.org/10.18653/v1/D19-1418} {Don't take the easy way out: Ensemble based methods for avoiding known dataset biases}.
\newblock In \emph{Proceedings of the 2019 Conference on Empirical Methods in Natural Language Processing and the 9th International Joint Conference on Natural Language Processing (EMNLP-IJCNLP)}, pages 4069--4082, Hong Kong, China. Association for Computational Linguistics.

\bibitem[{Creager et~al.(2021)Creager, Jacobsen, and Zemel}]{pmlr-v139-creager21a}
Elliot Creager, Joern-Henrik Jacobsen, and Richard Zemel. 2021.
\newblock \href {https://proceedings.mlr.press/v139/creager21a.html} {Environment inference for invariant learning}.
\newblock In \emph{Proceedings of the 38th International Conference on Machine Learning}, volume 139 of \emph{Proceedings of Machine Learning Research}, pages 2189--2200. PMLR.

\bibitem[{Dempster et~al.(1977)Dempster, Laird, and Rubin}]{dempster1977}
Arthur~P. Dempster, Nan~M. Laird, and Donald~B. Rubin. 1977.
\newblock \href {https://doi.org/10.1111/j.2517-6161.1977.tb01600.x} {Maximum likelihood from incomplete data via the em algorithm}.
\newblock \emph{Journal of the Royal Statistical Society: Series B (Methodological)}, 39(1):1--38.

\bibitem[{Dong et~al.(2024)Dong, Li, Dai, Zheng, Ma, Li, Xia, Xu, Wu, Chang, Sun, Li, and Sui}]{dong-etal-2024-survey}
Qingxiu Dong, Lei Li, Damai Dai, Ce~Zheng, Jingyuan Ma, Rui Li, Heming Xia, Jingjing Xu, Zhiyong Wu, Baobao Chang, Xu~Sun, Lei Li, and Zhifang Sui. 2024.
\newblock \href {https://doi.org/10.18653/v1/2024.emnlp-main.64} {A survey on in-context learning}.
\newblock In \emph{Proceedings of the 2024 Conference on Empirical Methods in Natural Language Processing}, pages 1107--1128, Miami, Florida, USA. Association for Computational Linguistics.

\bibitem[{Geva et~al.(2019)Geva, Goldberg, and Berant}]{geva-etal-2019-modeling}
Mor Geva, Yoav Goldberg, and Jonathan Berant. 2019.
\newblock \href {https://doi.org/10.18653/v1/D19-1107} {Are we modeling the task or the annotator? an investigation of annotator bias in natural language understanding datasets}.
\newblock In \emph{Proceedings of the 2019 Conference on Empirical Methods in Natural Language Processing and the 9th International Joint Conference on Natural Language Processing (EMNLP-IJCNLP)}, pages 1161--1166, Hong Kong, China. Association for Computational Linguistics.

\bibitem[{Guo et~al.(2025)Guo, Yang, Zhang, Song, Zhang, Xu, Zhu, Ma, Wang, Bi et~al.}]{guo2025deepseek}
Daya Guo, Dejian Yang, Haowei Zhang, Junxiao Song, Ruoyu Zhang, Runxin Xu, Qihao Zhu, Shirong Ma, Peiyi Wang, Xiao Bi, and 1 others. 2025.
\newblock \href {https://arxiv.org/abs/2501.12948} {Deepseek-r1: Incentivizing reasoning capability in llms via reinforcement learning}.
\newblock \emph{arXiv preprint arXiv:2501.12948v1}.

\bibitem[{Gupta et~al.(2023)Gupta, Gardner, and Singh}]{gupta-etal-2023-coverage}
Shivanshu Gupta, Matt Gardner, and Sameer Singh. 2023.
\newblock \href {https://doi.org/10.18653/v1/2023.findings-emnlp.930} {Coverage-based example selection for in-context learning}.
\newblock In \emph{Findings of the Association for Computational Linguistics: EMNLP 2023}, pages 13924--13950, Singapore. Association for Computational Linguistics.

\bibitem[{Gururangan et~al.(2018)Gururangan, Swayamdipta, Levy, Schwartz, Bowman, and Smith}]{gururangan-etal-2018-annotation}
Suchin Gururangan, Swabha Swayamdipta, Omer Levy, Roy Schwartz, Samuel Bowman, and Noah~A. Smith. 2018.
\newblock \href {https://doi.org/10.18653/v1/N18-2017} {Annotation artifacts in natural language inference data}.
\newblock In \emph{Proceedings of the 2018 Conference of the North {A}merican Chapter of the Association for Computational Linguistics: Human Language Technologies, Volume 2 (Short Papers)}, pages 107--112, New Orleans, Louisiana. Association for Computational Linguistics.

\bibitem[{He et~al.(2019)He, Zha, and Wang}]{he-etal-2019-unlearn}
He~He, Sheng Zha, and Haohan Wang. 2019.
\newblock \href {https://doi.org/10.18653/v1/D19-6115} {Unlearn dataset bias in natural language inference by fitting the residual}.
\newblock In \emph{Proceedings of the 2nd Workshop on Deep Learning Approaches for Low-Resource NLP (DeepLo 2019)}, pages 132--142, Hong Kong, China. Association for Computational Linguistics.

\bibitem[{He et~al.(2024)He, Wu, Camburu, Minervini, and Stenetorp}]{he-etal-2024-using}
Xuanli He, Yuxiang Wu, Oana-Maria Camburu, Pasquale Minervini, and Pontus Stenetorp. 2024.
\newblock \href {https://doi.org/10.18653/v1/2024.acl-long.728} {Using natural language explanations to improve robustness of in-context learning}.
\newblock In \emph{Proceedings of the 62nd Annual Meeting of the Association for Computational Linguistics (Volume 1: Long Papers)}, pages 13477--13499, Bangkok, Thailand. Association for Computational Linguistics.

\bibitem[{Honda et~al.(2024)Honda, Oka, Zhang, and Mita}]{honda-etal-2024-eliminate}
Ukyo Honda, Tatsushi Oka, Peinan Zhang, and Masato Mita. 2024.
\newblock \href {https://doi.org/10.1162/tacl_a_00701} {Not eliminate but aggregate: Post-hoc control over mixture-of-experts to address shortcut shifts in natural language understanding}.
\newblock \emph{Transactions of the Association for Computational Linguistics}, 12:1268--1289.

\bibitem[{Iyer et~al.(2017)Iyer, Dandekar, and Csernai}]{qqp2017}
Shankar Iyer, Nikhil Dandekar, and Kornel Csernai. 2017.
\newblock \href {https://quoradata.quora.com/First-Quora-Dataset-Release-Question-Pairs} {First quora dataset release: Question pairs}.
\newblock Accessed on Feb. 9, 2025.

\bibitem[{Jang et~al.(2024)Jang, Jang, Kweon, Jeon, and Yu}]{jang-etal-2024-rectifying}
Joonwon Jang, Sanghwan Jang, Wonbin Kweon, Minjin Jeon, and Hwanjo Yu. 2024.
\newblock \href {https://doi.org/10.18653/v1/2024.naacl-long.242} {Rectifying demonstration shortcut in in-context learning}.
\newblock In \emph{Proceedings of the 2024 Conference of the North American Chapter of the Association for Computational Linguistics: Human Language Technologies (Volume 1: Long Papers)}, pages 4294--4321, Mexico City, Mexico. Association for Computational Linguistics.

\bibitem[{Jiang and Marneffe(2022)}]{10.1162/tacl_a_00523}
Nan-Jiang Jiang and Marie-Catherine~de Marneffe. 2022.
\newblock \href {https://doi.org/10.1162/tacl_a_00523} {Investigating reasons for disagreement in natural language inference}.
\newblock \emph{Transactions of the Association for Computational Linguistics}, 10:1357--1374.

\bibitem[{Karimi~Mahabadi et~al.(2020)Karimi~Mahabadi, Belinkov, and Henderson}]{karimi-mahabadi-etal-2020-end}
Rabeeh Karimi~Mahabadi, Yonatan Belinkov, and James Henderson. 2020.
\newblock \href {https://doi.org/10.18653/v1/2020.acl-main.769} {End-to-end bias mitigation by modelling biases in corpora}.
\newblock In \emph{Proceedings of the 58th Annual Meeting of the Association for Computational Linguistics}, pages 8706--8716, Online. Association for Computational Linguistics.

\bibitem[{Kavumba et~al.(2023)Kavumba, Brassard, Heinzerling, and Inui}]{kavumba-etal-2023-prompting}
Pride Kavumba, Ana Brassard, Benjamin Heinzerling, and Kentaro Inui. 2023.
\newblock \href {https://doi.org/10.18653/v1/2023.findings-eacl.162} {Prompting for explanations improves adversarial {NLI}. is this true? {\{}yes{\}} it is {\{}true{\}} because {\{}it weakens superficial cues{\}}}.
\newblock In \emph{Findings of the Association for Computational Linguistics: EACL 2023}, pages 2165--2180, Dubrovnik, Croatia. Association for Computational Linguistics.

\bibitem[{Kojima et~al.(2022)Kojima, Gu, Reid, Matsuo, and Iwasawa}]{kojima2022large}
Takeshi Kojima, Shixiang~(Shane) Gu, Machel Reid, Yutaka Matsuo, and Yusuke Iwasawa. 2022.
\newblock \href {https://proceedings.neurips.cc/paper_files/paper/2022/file/8bb0d291acd4acf06ef112099c16f326-Paper-Conference.pdf} {Large language models are zero-shot reasoners}.
\newblock In \emph{Advances in Neural Information Processing Systems}, volume~35, pages 22199--22213. Curran Associates, Inc.

\bibitem[{Levy et~al.(2023)Levy, Bogin, and Berant}]{levy-etal-2023-diverse}
Itay Levy, Ben Bogin, and Jonathan Berant. 2023.
\newblock \href {https://doi.org/10.18653/v1/2023.acl-long.78} {Diverse demonstrations improve in-context compositional generalization}.
\newblock In \emph{Proceedings of the 61st Annual Meeting of the Association for Computational Linguistics (Volume 1: Long Papers)}, pages 1401--1422, Toronto, Canada. Association for Computational Linguistics.

\bibitem[{Li et~al.(2023)Li, Wang, Feng, Cao, Zhang, and Chua}]{li-etal-2023-robust}
Moxin Li, Wenjie Wang, Fuli Feng, Yixin Cao, Jizhi Zhang, and Tat-Seng Chua. 2023.
\newblock \href {https://doi.org/10.18653/v1/2023.emnlp-main.95} {Robust prompt optimization for large language models against distribution shifts}.
\newblock In \emph{Proceedings of the 2023 Conference on Empirical Methods in Natural Language Processing}, pages 1539--1554, Singapore. Association for Computational Linguistics.

\bibitem[{Liu et~al.(2021)Liu, Haghgoo, Chen, Raghunathan, Koh, Sagawa, Liang, and Finn}]{liu2021just}
Evan~Z Liu, Behzad Haghgoo, Annie~S Chen, Aditi Raghunathan, Pang~Wei Koh, Shiori Sagawa, Percy Liang, and Chelsea Finn. 2021.
\newblock \href {https://proceedings.mlr.press/v139/liu21f.html} {Just train twice: Improving group robustness without training group information}.
\newblock In \emph{Proceedings of the 38th International Conference on Machine Learning}, volume 139 of \emph{Proceedings of Machine Learning Research}, pages 6781--6792. PMLR.

\bibitem[{Liu et~al.(2022)Liu, Shen, Zhang, Dolan, Carin, and Chen}]{liu-etal-2022-makes}
Jiachang Liu, Dinghan Shen, Yizhe Zhang, Bill Dolan, Lawrence Carin, and Weizhu Chen. 2022.
\newblock \href {https://doi.org/10.18653/v1/2022.deelio-1.10} {What makes good in-context examples for {GPT}-3?}
\newblock In \emph{Proceedings of Deep Learning Inside Out (DeeLIO 2022): The 3rd Workshop on Knowledge Extraction and Integration for Deep Learning Architectures}, pages 100--114, Dublin, Ireland and Online. Association for Computational Linguistics.

\bibitem[{Liu et~al.(2020{\natexlab{a}})Liu, Xin, Chang, and Sui}]{liu-etal-2020-hyponli}
Tianyu Liu, Zheng Xin, Baobao Chang, and Zhifang Sui. 2020{\natexlab{a}}.
\newblock \href {https://aclanthology.org/2020.lrec-1.846/} {{H}ypo{NLI}: Exploring the artificial patterns of hypothesis-only bias in natural language inference}.
\newblock In \emph{Proceedings of the Twelfth Language Resources and Evaluation Conference}, pages 6852--6860, Marseille, France. European Language Resources Association.

\bibitem[{Liu et~al.(2020{\natexlab{b}})Liu, Xin, Ding, Chang, and Sui}]{liu-etal-2020-empirical}
Tianyu Liu, Zheng Xin, Xiaoan Ding, Baobao Chang, and Zhifang Sui. 2020{\natexlab{b}}.
\newblock \href {https://doi.org/10.18653/v1/2020.conll-1.48} {An empirical study on model-agnostic debiasing strategies for robust natural language inference}.
\newblock In \emph{Proceedings of the 24th Conference on Computational Natural Language Learning}, pages 596--608, Online. Association for Computational Linguistics.

\bibitem[{Luce(1959)}]{luce1959}
R.~Duncan Luce. 1959.
\newblock \emph{Individual Choice Behavior: A Theoretical Analysis}.
\newblock Wiley, New York.
\newblock Reprinted by Dover Publications, 2005.

\bibitem[{Ludan et~al.(2023)Ludan, Meng, Nguyen, Shah, Lyu, Apidianaki, and Callison-Burch}]{ludan-etal-2023-explanation}
Josh~Magnus Ludan, Yixuan Meng, Tai Nguyen, Saurabh Shah, Qing Lyu, Marianna Apidianaki, and Chris Callison-Burch. 2023.
\newblock \href {https://doi.org/10.18653/v1/2023.acl-long.242} {Explanation-based finetuning makes models more robust to spurious cues}.
\newblock In \emph{Proceedings of the 61st Annual Meeting of the Association for Computational Linguistics (Volume 1: Long Papers)}, pages 4420--4441, Toronto, Canada. Association for Computational Linguistics.

\bibitem[{Marschak(1960)}]{marschak1960}
Jacob Marschak. 1960.
\newblock Binary choice constraints and random utility indicators.
\newblock \emph{Stanford Research Memorandum}.

\bibitem[{McCoy et~al.(2019)McCoy, Pavlick, and Linzen}]{mccoy-etal-2019-right}
Tom McCoy, Ellie Pavlick, and Tal Linzen. 2019.
\newblock \href {https://doi.org/10.18653/v1/P19-1334} {Right for the wrong reasons: Diagnosing syntactic heuristics in natural language inference}.
\newblock In \emph{Proceedings of the 57th Annual Meeting of the Association for Computational Linguistics}, pages 3428--3448, Florence, Italy. Association for Computational Linguistics.

\bibitem[{McFadden(1973)}]{mcfadden1973}
Daniel McFadden. 1973.
\newblock Conditional logit analysis of qualitative choice behavior.
\newblock In P.~Zarembka, editor, \emph{Frontiers in Econometrics}, pages 105--142. Academic Press.

\bibitem[{Min et~al.(2022)Min, Lyu, Holtzman, Artetxe, Lewis, Hajishirzi, and Zettlemoyer}]{min-etal-2022-rethinking}
Sewon Min, Xinxi Lyu, Ari Holtzman, Mikel Artetxe, Mike Lewis, Hannaneh Hajishirzi, and Luke Zettlemoyer. 2022.
\newblock \href {https://doi.org/10.18653/v1/2022.emnlp-main.759} {Rethinking the role of demonstrations: What makes in-context learning work?}
\newblock In \emph{Proceedings of the 2022 Conference on Empirical Methods in Natural Language Processing}, pages 11048--11064, Abu Dhabi, United Arab Emirates. Association for Computational Linguistics.

\bibitem[{Mueller et~al.(2024)Mueller, Webson, Petty, and Linzen}]{mueller-etal-2024-context}
Aaron Mueller, Albert Webson, Jackson Petty, and Tal Linzen. 2024.
\newblock \href {https://doi.org/10.18653/v1/2024.naacl-long.267} {In-context learning generalizes, but not always robustly: The case of syntax}.
\newblock In \emph{Proceedings of the 2024 Conference of the North American Chapter of the Association for Computational Linguistics: Human Language Technologies (Volume 1: Long Papers)}, pages 4761--4779, Mexico City, Mexico. Association for Computational Linguistics.

\bibitem[{Naik et~al.(2018)Naik, Ravichander, Sadeh, Rose, and Neubig}]{naik-etal-2018-stress}
Aakanksha Naik, Abhilasha Ravichander, Norman Sadeh, Carolyn Rose, and Graham Neubig. 2018.
\newblock \href {https://aclanthology.org/C18-1198/} {Stress test evaluation for natural language inference}.
\newblock In \emph{Proceedings of the 27th International Conference on Computational Linguistics}, pages 2340--2353, Santa Fe, New Mexico, USA. Association for Computational Linguistics.

\bibitem[{Nie et~al.(2020{\natexlab{a}})Nie, Williams, Dinan, Bansal, Weston, and Kiela}]{nie-etal-2020-adversarial}
Yixin Nie, Adina Williams, Emily Dinan, Mohit Bansal, Jason Weston, and Douwe Kiela. 2020{\natexlab{a}}.
\newblock \href {https://doi.org/10.18653/v1/2020.acl-main.441} {Adversarial {NLI}: A new benchmark for natural language understanding}.
\newblock In \emph{Proceedings of the 58th Annual Meeting of the Association for Computational Linguistics}, pages 4885--4901, Online. Association for Computational Linguistics.

\bibitem[{Nie et~al.(2020{\natexlab{b}})Nie, Zhou, and Bansal}]{nie-etal-2020-learn}
Yixin Nie, Xiang Zhou, and Mohit Bansal. 2020{\natexlab{b}}.
\newblock \href {https://doi.org/10.18653/v1/2020.emnlp-main.734} {What can we learn from collective human opinions on natural language inference data?}
\newblock In \emph{Proceedings of the 2020 Conference on Empirical Methods in Natural Language Processing (EMNLP)}, pages 9131--9143, Online. Association for Computational Linguistics.

\bibitem[{Pavlick and Kwiatkowski(2019)}]{10.1162/tacl_a_00293}
Ellie Pavlick and Tom Kwiatkowski. 2019.
\newblock \href {https://doi.org/10.1162/tacl_a_00293} {Inherent disagreements in human textual inferences}.
\newblock \emph{Transactions of the Association for Computational Linguistics}, 7:677--694.

\bibitem[{Radford et~al.(2019)Radford, Wu, Child, Luan, Amodei, and Sutskever}]{radford2019language}
Alec Radford, Jeff Wu, Rewon Child, David Luan, Dario Amodei, and Ilya Sutskever. 2019.
\newblock Language models are unsupervised multitask learners.

\bibitem[{Sagawa et~al.(2020)Sagawa, Koh, Hashimoto, and Liang}]{sagawa2020distributionally}
Shiori Sagawa, Pang~Wei Koh, Tatsunori~B. Hashimoto, and Percy Liang. 2020.
\newblock \href {https://openreview.net/forum?id=ryxGuJrFvS} {Distributionally robust neural networks}.
\newblock In \emph{International Conference on Learning Representations}.

\bibitem[{Santurkar et~al.(2023)Santurkar, Durmus, Ladhak, Lee, Liang, and Hashimoto}]{pmlr-v202-santurkar23a}
Shibani Santurkar, Esin Durmus, Faisal Ladhak, Cinoo Lee, Percy Liang, and Tatsunori Hashimoto. 2023.
\newblock \href {https://proceedings.mlr.press/v202/santurkar23a.html} {Whose opinions do language models reflect?}
\newblock In \emph{Proceedings of the 40th International Conference on Machine Learning}, volume 202 of \emph{Proceedings of Machine Learning Research}, pages 29971--30004. PMLR.

\bibitem[{Siska et~al.(2024)Siska, Marazopoulou, Ailem, and Bono}]{siska-etal-2024-examining}
Charlotte Siska, Katerina Marazopoulou, Melissa Ailem, and James Bono. 2024.
\newblock \href {https://doi.org/10.18653/v1/2024.acl-long.560} {Examining the robustness of {LLM} evaluation to the distributional assumptions of benchmarks}.
\newblock In \emph{Proceedings of the 62nd Annual Meeting of the Association for Computational Linguistics (Volume 1: Long Papers)}, pages 10406--10421, Bangkok, Thailand. Association for Computational Linguistics.

\bibitem[{Stacey et~al.(2022)Stacey, Belinkov, and Rei}]{stacey2022supervising}
Joe Stacey, Yonatan Belinkov, and Marek Rei. 2022.
\newblock \href {https://doi.org/10.1609/aaai.v36i10.21386} {Supervising model attention with human explanations for robust natural language inference}.
\newblock \emph{Proceedings of the AAAI Conference on Artificial Intelligence}, 36(10):11349--11357.

\bibitem[{Sun et~al.(2024)Sun, Xiao, Li, Ji, Chen, and Zhang}]{sun-etal-2024-exploring}
Zechen Sun, Yisheng Xiao, Juntao Li, Yixin Ji, Wenliang Chen, and Min Zhang. 2024.
\newblock \href {https://aclanthology.org/2024.lrec-main.602/} {Exploring and mitigating shortcut learning for generative large language models}.
\newblock In \emph{Proceedings of the 2024 Joint International Conference on Computational Linguistics, Language Resources and Evaluation (LREC-COLING 2024)}, pages 6883--6893, Torino, Italia. ELRA and ICCL.

\bibitem[{Tang et~al.(2023)Tang, Kong, Huang, and Xue}]{tang-etal-2023-large}
Ruixiang Tang, Dehan Kong, Longtao Huang, and Hui Xue. 2023.
\newblock \href {https://doi.org/10.18653/v1/2023.findings-acl.284} {Large language models can be lazy learners: Analyze shortcuts in in-context learning}.
\newblock In \emph{Findings of the Association for Computational Linguistics: ACL 2023}, pages 4645--4657, Toronto, Canada. Association for Computational Linguistics.

\bibitem[{Thurstone(1927)}]{thurstone1927}
L.~L. Thurstone. 1927.
\newblock \href {https://doi.org/10.1037/h0070288} {A law of comparative judgment}.
\newblock \emph{Psychological Review}, 34(4):273--286.

\bibitem[{Truong et~al.(2022)Truong, Otmakhova, Baldwin, Cohn, Lau, and Verspoor}]{truong-etal-2022-another}
Thinh~Hung Truong, Yulia Otmakhova, Timothy Baldwin, Trevor Cohn, Jey~Han Lau, and Karin Verspoor. 2022.
\newblock \href {https://doi.org/10.18653/v1/2022.aacl-main.65} {Not another negation benchmark: The {N}a{N}-{NLI} test suite for sub-clausal negation}.
\newblock In \emph{Proceedings of the 2nd Conference of the Asia-Pacific Chapter of the Association for Computational Linguistics and the 12th International Joint Conference on Natural Language Processing (Volume 1: Long Papers)}, pages 883--894, Online only. Association for Computational Linguistics.

\bibitem[{Wang et~al.(2023{\natexlab{a}})Wang, HU, Hou, Chen, Zheng, Wang, Yang, Ye, Huang, Geng, Jiao, Zhang, and Xie}]{wang2023on}
Jindong Wang, Xixu HU, Wenxin Hou, Hao Chen, Runkai Zheng, Yidong Wang, Linyi Yang, Wei Ye, Haojun Huang, Xiubo Geng, Binxing Jiao, Yue Zhang, and Xing Xie. 2023{\natexlab{a}}.
\newblock \href {https://openreview.net/forum?id=uw6HSkgoM29} {On the robustness of chat{GPT}: An adversarial and out-of-distribution perspective}.
\newblock In \emph{ICLR 2023 Workshop on Trustworthy and Reliable Large-Scale Machine Learning Models}.

\bibitem[{Wang et~al.(2023{\natexlab{b}})Wang, Liu, Park, Jiang, Zheng, Wu, Chen, and Xiao}]{wang2023adversarial}
Jiongxiao Wang, Zichen Liu, Keun~Hee Park, Zhuojun Jiang, Zhaoheng Zheng, Zhuofeng Wu, Muhao Chen, and Chaowei Xiao. 2023{\natexlab{b}}.
\newblock \href {https://arxiv.org/abs/2305.14950} {Adversarial demonstration attacks on large language models}.
\newblock \emph{arXiv preprint arXiv:2305.14950v2}.

\bibitem[{Wang et~al.(2023{\natexlab{c}})Wang, Wei, Schuurmans, Le, Chi, Narang, Chowdhery, and Zhou}]{wang2023selfconsistency}
Xuezhi Wang, Jason Wei, Dale Schuurmans, Quoc~V Le, Ed~H. Chi, Sharan Narang, Aakanksha Chowdhery, and Denny Zhou. 2023{\natexlab{c}}.
\newblock \href {https://openreview.net/forum?id=1PL1NIMMrw} {Self-consistency improves chain of thought reasoning in language models}.
\newblock In \emph{The Eleventh International Conference on Learning Representations}.

\bibitem[{Wei et~al.(2022)Wei, Wang, Schuurmans, Bosma, ichter, Xia, Chi, Le, and Zhou}]{wei2022chain}
Jason Wei, Xuezhi Wang, Dale Schuurmans, Maarten Bosma, brian ichter, Fei Xia, Ed~Chi, Quoc~V Le, and Denny Zhou. 2022.
\newblock \href {https://proceedings.neurips.cc/paper_files/paper/2022/file/9d5609613524ecf4f15af0f7b31abca4-Paper-Conference.pdf} {Chain-of-thought prompting elicits reasoning in large language models}.
\newblock In \emph{Advances in Neural Information Processing Systems}, volume~35, pages 24824--24837. Curran Associates, Inc.

\bibitem[{Wolf et~al.(2020)Wolf, Debut, Sanh, Chaumond, Delangue, Moi, Cistac, Rault, Louf, Funtowicz, Davison, Shleifer, von Platen, Ma, Jernite, Plu, Xu, Le~Scao, Gugger, Drame, Lhoest, and Rush}]{wolf-etal-2020-transformers}
Thomas Wolf, Lysandre Debut, Victor Sanh, Julien Chaumond, Clement Delangue, Anthony Moi, Pierric Cistac, Tim Rault, Remi Louf, Morgan Funtowicz, Joe Davison, Sam Shleifer, Patrick von Platen, Clara Ma, Yacine Jernite, Julien Plu, Canwen Xu, Teven Le~Scao, Sylvain Gugger, and 3 others. 2020.
\newblock \href {https://doi.org/10.18653/v1/2020.emnlp-demos.6} {Transformers: State-of-the-art natural language processing}.
\newblock In \emph{Proceedings of the 2020 Conference on Empirical Methods in Natural Language Processing: System Demonstrations}, pages 38--45, Online. Association for Computational Linguistics.

\bibitem[{Yang et~al.(2023)Yang, Zhang, Katabi, and Ghassemi}]{pmlr-v202-yang23s}
Yuzhe Yang, Haoran Zhang, Dina Katabi, and Marzyeh Ghassemi. 2023.
\newblock \href {https://proceedings.mlr.press/v202/yang23s.html} {Change is hard: A closer look at subpopulation shift}.
\newblock In \emph{Proceedings of the 40th International Conference on Machine Learning}, volume 202 of \emph{Proceedings of Machine Learning Research}, pages 39584--39622. PMLR.

\bibitem[{Ye et~al.(2023)Ye, Wu, Feng, Yu, and Kong}]{pmlr-v202-ye23c}
Jiacheng Ye, Zhiyong Wu, Jiangtao Feng, Tao Yu, and Lingpeng Kong. 2023.
\newblock \href {https://proceedings.mlr.press/v202/ye23c.html} {Compositional exemplars for in-context learning}.
\newblock In \emph{Proceedings of the 40th International Conference on Machine Learning}, volume 202 of \emph{Proceedings of Machine Learning Research}, pages 39818--39833. PMLR.

\bibitem[{Yuan et~al.(2024)Yuan, Zhao, Zhang, Zheng, and Liu}]{yuan-etal-2024-llms}
Yu~Yuan, Lili Zhao, Kai Zhang, Guangting Zheng, and Qi~Liu. 2024.
\newblock \href {https://doi.org/10.18653/v1/2024.emnlp-main.679} {Do {LLM}s overcome shortcut learning? an evaluation of shortcut challenges in large language models}.
\newblock In \emph{Proceedings of the 2024 Conference on Empirical Methods in Natural Language Processing}, pages 12188--12200, Miami, Florida, USA. Association for Computational Linguistics.

\bibitem[{Zhang et~al.(2020)Zhang, Kishore, Wu, Weinberger, and Artzi}]{Zhang2020BERTScore:}
Tianyi Zhang, Varsha Kishore, Felix Wu, Kilian~Q. Weinberger, and Yoav Artzi. 2020.
\newblock \href {https://openreview.net/forum?id=SkeHuCVFDr} {Bertscore: Evaluating text generation with bert}.
\newblock In \emph{International Conference on Learning Representations}.

\bibitem[{Zhang et~al.(2019)Zhang, Baldridge, and He}]{zhang-etal-2019-paws}
Yuan Zhang, Jason Baldridge, and Luheng He. 2019.
\newblock \href {https://doi.org/10.18653/v1/N19-1131} {{PAWS}: Paraphrase adversaries from word scrambling}.
\newblock In \emph{Proceedings of the 2019 Conference of the North {A}merican Chapter of the Association for Computational Linguistics: Human Language Technologies, Volume 1 (Long and Short Papers)}, pages 1298--1308, Minneapolis, Minnesota. Association for Computational Linguistics.

\bibitem[{Zhou et~al.(2024)Zhou, Xu, Liu, An, Ai, and Huang}]{zhou-etal-2024-explore}
Yuhang Zhou, Paiheng Xu, Xiaoyu Liu, Bang An, Wei Ai, and Furong Huang. 2024.
\newblock \href {https://doi.org/10.18653/v1/2024.acl-long.28} {Explore spurious correlations at the concept level in language models for text classification}.
\newblock In \emph{Proceedings of the 62nd Annual Meeting of the Association for Computational Linguistics (Volume 1: Long Papers)}, pages 478--492, Bangkok, Thailand. Association for Computational Linguistics.

\end{thebibliography}

\clearpage
\appendix

\section{Algorithms}
\label{sec:appendix:algo}
We present the algorithms for ICL and X-ICL in Algorithms~\ref{alg:icl} and~\ref{alg:x-icl}, respectively.

\begin{algorithm}[tb]
\caption{ICL}
\label{alg:icl}
\begin{algorithmic}[1]
\INPUT \ $\D_{n}$
\STATE \textbf{Estimation:}
\STATE \quad Data: $\D_{n}$
\STATE \quad Model: $p(y|x)$
\STATE \quad Estimator: 
$\hat{p}(y|x)$
\STATE \textbf{Inference:} 
\STATE \quad Test input: $x'$
\STATE \quad Compute 
    $\{ \hat{p}(y'| x'): \ y' \in \Y \}$
\OUTPUT \ Classification 
\STATE \quad
$y^{\ast} = \argmax_{y' \in \mathcal{Y}} \hat{p}(y'| x')$
\end{algorithmic}
\end{algorithm}

\section{Dataset Details}
\label{sec:appendix:datasets}
\looseness=-1
HANS is characterized by a shift in the correlation between the labels and the word overlap of the premise and the hypothesis.
In HANS, predictions based solely on word overlap can achieve only chance-level accuracy.
NAN is constructed by combining multiple linguistic rules related to negation, which gives rise to rarely observed patterns in the standard NLI datasets.
PISP identifies specific patterns occurring in the hypotheses of each label in NLI datasets and constructs a dataset consisting of examples that do not adhere to the observed pattern--label co-occurrence.
ST synthesizes adversarial examples targeting multiple patterns that correlate with labels, including word overlap, negation, sentence length, etc.
ANLI consists of three distinct splits, R1, R2, and R3, each generated through a different round of adversarial data collection conducted by human annotators.
In our study, we specifically used the hard subsets,\footnote{\url{https://github.com/xlhex/acl2024_xicl/tree/main/data/testset}} in which the assigned labels were verified by a third human annotator to ensure reliability.\footnote{\url{https://github.com/facebookresearch/anli/blob/main/mds/verifier_labels.md}}
PAWS is a dataset in which the correlation between the labels and word overlap is adversarially shifted from QQP.

\section{Decoding Configurations}
\label{sec:appendix:models}
We used Azure OpenAI API to run GPT-4o and Gemini API for the Gemini models.
The Azure content filters were set to the least restrictive setting.
For the open-source models, we used Hugging Face Transformers library \citep{wolf-etal-2020-transformers}.\footnote{\url{https://github.com/huggingface/transformers}}
The models were loaded with the default configurations and run on a single NVIDIA A100 GPU with 40GB of memory.

To maximize reproducibility, we set the decoding temperature to zero for the closed-source models: GPT-4o, Gemini-1.5-Pro, and Gemini-2.0-Flash.
Note that deterministic decoding methods are not available for these models.
For the open-source models, Phi-4 and DeepSeek-R1-8B, we employed greedy decoding.

During preliminary experiments, we observed that the open-source models exhibited weaknesses in adhering to the expected output format, often generating responses that were difficult to parse.
To mitigate this issue, we incorporated the following system prompt to explicitly enforce adherence to the required output format: 
\begin{quote}
\emph{Answer the question by following the provided examples. Ensure that your response ends with Label: and your final answer.}
\end{quote}

\begin{algorithm}[tb]
\caption{X-ICL}
\label{alg:x-icl}
\begin{algorithmic}[1]
\INPUT \ $\D_{n}$ and $\Dmeta_{m}$
\FOR{$i = 1, \dots, n $}
    \STATE 
    Generate a latent variable
    $r_{y_i}$
    for $x_{i}$: \\
    \hspace{1.5cm} $r_{y_i} \sim \tilde{p}(r_{y_i}| y_{i}, x_{i})$
\ENDFOR
\STATE \textbf{Estimation:}
\STATE \quad Augumented data: $\{(x_i,r_{y_i},y_i)\}_{i=1}^n$
\STATE \quad Model: $p(y, r_{y}|x)$
\STATE \quad Estimator: 
$\hat{p}(y, r_{y}|x) = \hat{p}(y| r_{y}, x) \hat{p}(r_{y}|x)$
\STATE \textbf{Inference:} 
\STATE \quad Test input: $x'$
\STATE \quad Draw $r_{y}' \sim \hat{p}(r_{y}|x')$
\STATE \quad Compute 
    $\{ \hat{p}(y'|r_{y}', x'): \ y' \in \Y \}$
\OUTPUT \ Classification 
\STATE \quad
$y^{\ast} = \argmax_{y' \in \mathcal{Y}} \hat{p}(y'| r_{y}', x')$
\end{algorithmic}
\end{algorithm}

\section{Comparison of Costs}
\label{sec:appendix:cost}

\begin{table}[t]
\centering
\begin{adjustbox}{max width=1.0\columnwidth}
\begin{tabular}{lcccc}
\toprule
 & \multicolumn{2}{c}{\textbf{Number of Tokens}} & \textbf{Wall-Clock Time} & \textbf{Expense} \\
 & Input & Output & Seconds & USD \\
\midrule
ICL & 366.3 & 3.3 & 0.38 & 0.0009 \\
X-ICL & 539.3 & 36.5 & 0.79 & 0.0017 \\
\rowcolor{backgray}
\xxicl{} & 1,240.3 & 120.3 & 1.83 & 0.0043 \\
\bottomrule
\end{tabular}
\end{adjustbox}
\caption{
Average costs per instance.
Wall-clock time is measured in seconds, and expense in US dollars.
}
\label{tab:costs}
\end{table}

We present the average costs per instance, including demonstrations prepended to each instance, based on experiments conducted with GPT-4o on the SNLI dataset using the same seed.
Here, an ``instance'' refers to a single data point; in NLI tasks, this is a pair consisting of a premise and a hypothesis.
We used OpenAI tiktoken with the \texttt{o200k\_base} encoding to count the tokens.\footnote{\url{https://github.com/openai/tiktoken}}
The expenses are calculated based on the number of tokens and the GPT-4o (\texttt{2024-08-06 Global}) pricing for the Azure OpenAI API as of May 2025.

Table~\ref{tab:costs} shows the results.
We observed that \xxicl{} incurs higher computational costs than ICL and X-ICL due to its more comprehensive reasoning process.
This is a limitation of our method, and reducing the costs remains future work.
However, we emphasize that cost efficiency is not the primary focus of our work.
Our main objective is to improve OOD robustness, and apart from the computational costs, \xxicl{} does not require any additional resources compared to X-ICL.

\section{Comparison with Other Non-Explanation-Based Methods}
\label{sec:appendix:expt_mixp}
\citet{sun-etal-2024-exploring} proposed simple methods to enhance the robustness of ICL.
Their first finding is that increasing the number of demonstrations improves robustness.
Additionally, they introduced a technique in which the original labels are replaced with random symbols, accompanied by a minimal description that maps these symbols to their corresponding labels.
This approach aims to mitigate spurious correlations between labels and target tasks that LLMs may learn.
Unlike other prior approaches, these methods offer a notable advantage in terms of ease of implementation: they only need to increase the number of demonstrations or replace the original labels with random symbols, without requiring any other modifications.

We evaluated the effectiveness of these methods in comparison to explanation-based approaches: X-ICL and \xxicl{}.
For the implementation of the first method, we concatenated all four sets of demonstrations utilized in the main experiment, resulting in a single run of 32-shot ICL.
For the second method, referred to as \emph{mixed prompts}, we replaced the original labels with random symbols and provided a minimal task description, following the format of \citet{sun-etal-2024-exploring}.
The task description we employed is as follows:
\begin{quote}
\emph{Given a premise and a hypothesis, if the premise entails the hypothesis, the answer is A4; if the hypothesis contradicts the premise, the answer is B6; and if the premise and hypothesis are unrelated, the answer is 7X.}
\end{quote}
The results are reported in Table~\ref{tab:ablation_not_reason}.
In our experiments using GPT-4o on the ANLI test sets, we did not observe any improvement over standard ICL when applying these methods. This suggests that the benefits of these techniques may not consistently generalize across different datasets and model architectures.
In contrast, explanation-based approaches have demonstrated effectiveness in a broader range of settings (see also Table~\ref{tab:main_results}), highlighting their robustness.

\begin{table}[t]
\centering
\begin{adjustbox}{max width=1.0\columnwidth}
\begin{tabular}{lccc}
\toprule
 & \textbf{ANLI$_{\textit{R1}}$} & \textbf{ANLI$_{\textit{R2}}$} & \textbf{ANLI$_{\textit{R3}}$} \\
\midrule
ICL (8-shot) & $70.67_{\pm 2.60}$ & $61.05_{\pm 2.23}$ & $61.58_{\pm 2.32}$ \\
\midrule
\citet{sun-etal-2024-exploring} \\
ICL (32-shot) & 70.00 & 60.06 & 63.16 \\
ICL + mixed prompts & $70.10_{\pm 1.15}$ & $57.93_{\pm 2.32}$ & $57.83_{\pm 2.27}$ \\
\midrule
X-ICL (8-shot) & $75.58_{\pm 0.92}$ & $63.87_{\pm 0.73}$ & $65.07_{\pm 2.23}$ \\
\rowcolor{backgray}
\xxicl{} (8-shot) & $\textbf{77.40}_{\pm 0.48}$ &  $\textbf{67.61}_{\pm 1.89}$ & $\textbf{67.70}_{\pm 1.63}$ \\
\bottomrule
\end{tabular}
\end{adjustbox}
\caption{
Comparison between explanation-based methods and those that do not incorporate explanations.
The results for GPT-4o are reported.
The notation follows that of Table~\ref{tab:main_results}.
}
\label{tab:ablation_not_reason}
\end{table}

\section{Prompt Examples}
\label{sec:appendix:prompts}
Figures~\ref{fig:prompt_xxicl} and \ref{fig:prompt_xxicl_qqp} present examples of the \xxicl{} prompt.
Figures~\ref{fig:xprompt_example_nli} and \ref{fig:xprompt_example_qqp} show the exact meta-prompts we used to generate explanations that were incorporated into the demonstrations.

\begin{figure*}[t]
\centering
\begin{tcolorbox}[title=\xxicl{} prompt for NLI, colback=gray!10, colframe=black!70, boxrule=0.5mm, arc=0.8mm, width=1.0\textwidth]
Premise: The Florida Scholastic Press Association (FSPA) is a not-for-profit organization founded in 1946. It is the scholastic press association for the state of Florida and its members consist of more than 300 student publications, online media teams and broadcast programs from the state. The main mission of the organization is to educate, train and support scholastic journalists and their advisers.\\
Hypothesis: The FSPA has over 300 members\\
Instruction: Explore the reasoning behind all the labels. Then, select the label that has the most valid reasoning.\\
Possible Reasoning for entailment: The premise states that FSPA's members consist of more than 300 student publications, online media teams, and broadcast programs, which entails that FSPA has over 300 members.\\
Possible Reasoning for neutral: The premise states that the FSPA's members consist of more than 300 student publications, online media teams, and broadcast programs, but it does not specify the exact number of members, so the hypothesis that the FSPA has over 300 members is not definitively supported or contradicted.\\
Possible Reasoning for contradiction: The premise states that the FSPA's members consist of more than 300 student publications, online media teams, and broadcast programs, not individual members, so the hypothesis incorrectly interprets the premise.\\
Label: entailment\\
\#\#\#\\
Premise: We Have Always Lived in the Castle is an upcoming American mystery thriller film directed by Stacie Passon and written by Passon and Mark Kruger, based on the 1962 novel of the same name by Shirley Jackson. The film stars Taissa Farmiga, Alexandra Daddario, Crispin Glover, and Sebastian Stan.\\
Hypothesis: We Have Always Lived in the Castle came out in 2018.\\
Instruction: Explore the reasoning behind all the labels. Then, select the label that has the most valid reasoning.\\
Possible Reasoning for entailment: The premise states that "We Have Always Lived in the Castle" is an upcoming film, which implies it was released after the premise was written, supporting the hypothesis that it came out in 2018.\\
Possible Reasoning for neutral: The premise states that the film is upcoming, but it does not provide information about its release year, so it is unclear if it came out in 2018.\\
Possible Reasoning for contradiction: The premise states that the film is upcoming, which contradicts the hypothesis that it was released in 2018.\\
Label: contradiction\\
\#\#\#\\
...\\
\#\#\#\\
Premise: \texttt{\{Test Premise\}}\\
Hypothesis: \texttt{\{Test Hypothesis\}}\\
Instruction: Explore the reasoning behind all the labels. Then, select the label that has the most valid reasoning.
\end{tcolorbox}
\caption{
An example of \xxicl{} prompt for NLI.
}
\label{fig:prompt_xxicl}
\end{figure*}
\begin{figure*}[t]
\centering
\begin{tcolorbox}[title=\xxicl{} prompt for paraphrase identification, colback=gray!10, colframe=black!70, boxrule=0.5mm, arc=0.8mm, width=1.0\textwidth]
Question 1: How can I be like Sheldon Cooper?\\
Question 2: How do I become like Sheldon Cooper?\\
Question: Do both questions ask the same thing?\\
Instruction: Explore the reasoning behind all the labels. Then, select the label that has the most valid reasoning.\\
Possible Reasoning for no: Although both questions seek advice about becoming like Sheldon Cooper, "can I be" asks about the possibility, while "how do I become" asks about the specific steps or process, making them technically different questions.\\
Possible Reasoning for yes: Both questions ask about ways to become/emulate the character Sheldon Cooper, using different but synonymous phrasing ("be like" vs. "become like").\\
Label: yes\\
\#\#\#\\
Question 1: Why can't some people think for themselves?\\
Question 2: Why don't people think for themselves?\\
Question: Do both questions ask the same thing?\\
Instruction: Explore the reasoning behind all the labels. Then, select the label that has the most valid reasoning.\\
Possible Reasoning for no: While similar, the first question asks about "some people" who are unable to think for themselves, while the second question asks about people in general who choose not to think for themselves. The scope and implication of inability versus choice are different.\\
Possible Reasoning for yes: Both questions ask about why people don't/can't engage in independent thinking, with slightly different wording but the same core meaning.\\
Label: no\\
\#\#\#\\
...\\
\#\#\#\\
Question 1: \texttt{\{Test Question 1\}}\\
Question 2: \texttt{\{Test Question 2\}}\\
Question: Do both questions ask the same thing?\\
Instruction: Explore the reasoning behind all the labels. Then, select the label that has the most valid reasoning.
\end{tcolorbox}
\caption{
An example of \xxicl{} prompt for paraphrase identification.
}
\label{fig:prompt_xxicl_qqp}
\end{figure*}
\begin{figure*}[t]
\centering
\begin{tcolorbox}[title=Meta-prompt for NLI, colback=gray!10, colframe=black!70, boxrule=0.5mm, arc=0.8mm, width=1.0\textwidth]
Assume that you're an expert working on natural language inference tasks. Given a premise, a hypothesis, and the corresponding label, please write a concise and precise reason to explain why the label is assigned to the example by following the provided examples:\\
\\
Premise: A boy peers out of an open window.\\
Hypothesis: The boy looks out the window.\\
Label: entailment\\
Reason: The boy peers out of a window, so the boy looks out the window.\\
\#\#\#\\
Premise: A kid doing a trick on a skateboard.\\
Hypothesis: The kid eating lunch inside the cafeteria.\\
Label: contradiction\\
Reason: The kid cannot be doing a trick and eating lunch at the same time\\
\#\#\#\\\
Premise: A man jumps off of his skateboard on the top of a cement ramp.\\
Hypothesis: a man jumps off a skateboard at the top of a ramp.\\
Label: neutral\\
Reason: A man can jump off a skateboard without being at the top of a ramp.\\
\#\#\#\\
Premise: \texttt{\{Demonstration Premise\}}\\
Hypothesis: \texttt{\{Demonstration Hypothesis\}}\\
Label: \texttt{\{Demonstration Label\}}
\end{tcolorbox}
\caption{Meta-prompt for generating explanations to be included in NLI demonstrations.
}
\label{fig:xprompt_example_nli}
\end{figure*}
\begin{figure*}[t]
\centering
\begin{tcolorbox}[title=Meta-prompt for paraphrase identification, colback=gray!10, colframe=black!70, boxrule=0.5mm, arc=0.8mm, width=1.0\textwidth]
Assume that you're an expert working on paraphrase identification tasks. Given questions 1 and 2 and the corresponding label, please write a concise and precise reason to explain why the label is assigned to the example by following the provided examples:\\
\\
Question 1: Does life get harder as you get older?\\
Question 2: Does life really get harder as you get older?\\
Question: Do both questions ask the same thing?\\
Label: yes\\
Reason: Both questions ask whether life does get harder as you get older.\\
\#\#\#\\
Question 1: What is the National nanotechnology initiative?\\
Question 2: What is the lead time for SSN4EGS411 board?\\
Question: Do both questions ask the same thing?\\
Label: no\\
Reason: completely different questions\\
\#\#\#\\
Question 1: \texttt{\{Demonstration Question 1\}}\\
Question 2: \texttt{\{Demonstration Question 2\}}\\
Question: Do both questions ask the same thing?\\
Label: \texttt{\{Demonstration Label\}}
\end{tcolorbox}
\caption{Meta-prompt for generating explanations to be included in paraphrase-identification demonstrations.
}
\label{fig:xprompt_example_qqp}
\end{figure*}

\section{Output Examples}
Figures~\ref{fig:output_pos_1}--\ref{fig:output_neg_4} illustrate the output examples of \xxicl{}.
See Section~\ref{sec:analysis} for a detailed discussion and analysis of these examples.

\label{sec:appendix:outputs}
\begin{figure*}[t]
\centering
\begin{tcolorbox}[title=Successful case of \xxicl{}, coltitle=black, colback=mygreen!30, colframe=mygreen, boxrule=0.5mm, arc=0.8mm, width=1.0\textwidth]
\small
\textbf{Premise:} In astrophysics, the Chandrasekhar equation is a dimensionless form of the Poisson equation for the density distribution of a spherically symmetric isothermal gas sphere subjected to its own gravitaional force, named after the Indian American astrophysicist Subrahmanyan Chandrasekhar. The equation reads\\
\textbf{Hypothesis:} Subrahmanyan Chandrasekhar was a very influential scientist. \\
\textbf{Label:} entailment\\

\textbf{Output of X-ICL}\\
Reason: The premise mentions that the Chandrasekhar equation is named after Subrahmanyan Chandrasekhar, suggesting his influence in the field of astrophysics. However, it does not provide enough information to definitively conclude the extent of his influence.\\
Label: neutral\\

\textbf{Output of \xxicl{}}\\
Possible Reasoning for entailment: The premise mentions that the Chandrasekhar equation is named after Subrahmanyan Chandrasekhar, which suggests he was influential enough to have an equation named after him.\\
Possible Reasoning for neutral: The premise provides information about the Chandrasekhar equation but does not explicitly state the level of influence Subrahmanyan Chandrasekhar had in the scientific community.\\
Possible Reasoning for contradiction: The premise does not provide any information that contradicts the hypothesis about Subrahmanyan Chandrasekhar's influence.\\
Label: entailment
\end{tcolorbox}
\caption{An example in which \xxicl{} provides the correct answer, whereas X-ICL fails.
\xxicl{} enabled the model to carefully examine the details of the input and to reason based on confidently inferred information.
}
\label{fig:output_pos_1}
\end{figure*}

\begin{figure*}[t]
\centering
\begin{tcolorbox}[title=Successful case of \xxicl{}, coltitle=black, colback=mygreen!30, colframe=mygreen, boxrule=0.5mm, arc=0.8mm, width=1.0\textwidth]
\small
\textbf{Premise:} David Carter (born November 27, 1953 in Vincennes, Indiana) is a former American football offensive guard and center who played 9 seasons in the National Football League for the Houston Oilers and New Orleans Saints. He played college football at Western Kentucky University.\\
\textbf{Hypothesis:} David Carter played professional football in the NFL until he was 52 years old.\\
\textbf{Label:} contradiction\\

\textbf{Output of X-ICL}\\
Reason: The premise does not provide information about the specific years David Carter played in the NFL, only that he played for 9 seasons. Without knowing the years he played, we cannot determine if he played until he was 52 years old.\\
Label: neutral\\

\textbf{Output of \xxicl{}}\\
Possible Reasoning for entailment: The premise provides information about David Carter's birthdate and his NFL career, but it does not specify the years he played, so it cannot confirm or deny the hypothesis.\\
Possible Reasoning for neutral: The premise states that David Carter played 9 seasons in the NFL, but it does not provide the specific years of his career, leaving his age at retirement uncertain.\\
Possible Reasoning for contradiction: Given that David Carter was born in 1953, if he played until he was 52 years old, he would have been playing in 2005. This is unlikely given the typical career length of NFL players and the fact that he played 9 seasons, suggesting he did not play until he was 52.\\
Label: contradiction
\end{tcolorbox}
\caption{An example in which \xxicl{} provides the correct answer, whereas X-ICL fails.
\xxicl{} enabled the model to carefully examine the details of the input and to reason based on confidently inferred information.
}
\label{fig:output_pos_2}
\end{figure*}
\begin{figure*}[t]
\centering
\begin{tcolorbox}[title=Successful case of \xxicl{}, coltitle=black, colback=mygreen!30, colframe=mygreen, boxrule=0.5mm, arc=0.8mm, width=1.0\textwidth]
\small
\textbf{Premise:} Thank You Happy Birthday is the second studio album by American rock band Cage the Elephant. It was released on January 11, 2011 to positive critical reception. The album was produced by Jay Joyce, who worked in the same capacity on the band's eponymous debut album.\\
\textbf{Hypothesis:} Cage the Elephant had positive reviews thanks to the work of Jay Joyce who produced it.\\
\textbf{Label:} neutral\\

\textbf{Output of X-ICL}\\
Reason: The premise states that "Thank You Happy Birthday" received positive critical reception and was produced by Jay Joyce. This supports the hypothesis that the positive reviews were due, at least in part, to Jay Joyce's production work.\\
Label: entailment\\

\textbf{Output of \xxicl{}}\\
Possible Reasoning for entailment: The premise states that "Thank You Happy Birthday" received positive critical reception and was produced by Jay Joyce, suggesting that his production work contributed to the positive reviews.\\
Possible Reasoning for neutral: The premise states that the album received positive reviews and was produced by Jay Joyce, but it does not explicitly link the positive reviews directly to his work, leaving room for other factors to have contributed to the reception.\\
Possible Reasoning for contradiction: The premise does not provide any information that contradicts the hypothesis, as it acknowledges both the positive reviews and Jay Joyce's role as producer.\\
Label: neutral
\end{tcolorbox}
\caption{An example in which \xxicl{} provides the correct answer, whereas X-ICL fails.
\xxicl{} prevented the model from making hasty inferences by drawing attention to the lack of decisive information in the given context.
}
\label{fig:output_pos_3}
\end{figure*}

\begin{figure*}[t]
\centering
\begin{tcolorbox}[title=Successful case of \xxicl{}, coltitle=black, colback=mygreen!30, colframe=mygreen, boxrule=0.5mm, arc=0.8mm, width=1.0\textwidth]
\small
\textbf{Premise:} The Best of David Bowie 1974/1979 is a compilation album by David Bowie released in 1998 (see 1998 in music). It follows "The Best of David Bowie 1969/1974" (1997) and includes material released between 1974–1979. This album was also included as the second disc of the compilation "The Platinum Collection" (2005/2006).\\
\textbf{Hypothesis:} David Bowie didn't only release an album in 1998 but also in 1979.\\
\textbf{Label:} neutral\\

\textbf{Output of X-ICL}\\
Reason: The premise mentions that "The Best of David Bowie 1974/1979" is a compilation album released in 1998, which includes material from 1974 to 1979. This implies that David Bowie released music during the years 1974 to 1979, including 1979, in addition to the compilation album released in 1998.\\
Label: entailment\\

\textbf{Output of \xxicl{}}\\
Possible Reasoning for entailment: The premise mentions that "The Best of David Bowie 1974/1979" is a compilation album released in 1998, which includes material from 1974 to 1979. This implies that David Bowie released music during those years, including 1979.\\
Possible Reasoning for neutral: The premise confirms the release of a compilation album in 1998 and includes material from 1974 to 1979, but it does not explicitly state that an album was released in 1979.\\
Possible Reasoning for contradiction: The premise does not explicitly state that David Bowie released an album in 1979, only that the compilation includes material from that year, so it does not directly support the hypothesis.\\
Label: neutral
\end{tcolorbox}
\caption{An example in which \xxicl{} provides the correct answer, whereas X-ICL fails.
\xxicl{} prevented the model from making hasty inferences by drawing attention to the lack of decisive information in the given context.
}
\label{fig:output_pos_4}
\end{figure*}
\begin{figure*}[t]
\centering
\begin{tcolorbox}[title=Failure case of \xxicl{}, coltitle=black, colback=myred!30, colframe=myred, boxrule=0.5mm, arc=0.8mm, width=1.0\textwidth]
\small
\textbf{Premise:} Garrett Smithley (born April 27, 1992) is an American professional stock car racing driver. He currently competes full-time in the NASCAR Xfinity Series, driving the No. 0 Chevrolet Camaro for JD Motorsports and the No. 40 Toyota Camry for MBM Motorsports. Smithley has also competed in the Camping World Truck Series and ARCA Racing Series.\\
\textbf{Hypothesis:} Garrett Smithley used to compete in the NASCAR Xfinity Series.\\
\textbf{Label:} contradiction\\

\textbf{Output of X-ICL}\\
Reason: The premise states that Garrett Smithley currently competes full-time in the NASCAR Xfinity Series, indicating that he is still active in the series, not that he used to compete in it.\\
Label: contradiction\\

\textbf{Output of \xxicl{}}\\
Possible Reasoning for entailment: The premise states that Garrett Smithley currently competes full-time in the NASCAR Xfinity Series, which implies that he used to compete in it as well.\\
Possible Reasoning for neutral: The premise provides information about Garrett Smithley's current participation in the NASCAR Xfinity Series, but it does not explicitly state his past involvement, leaving room for interpretation.\\
Possible Reasoning for contradiction: The premise states that Garrett Smithley currently competes in the NASCAR Xfinity Series, which does not contradict the hypothesis that he used to compete in it.\\
Label: entailment
\end{tcolorbox}
\caption{An example in which X-ICL provides the correct answer, whereas \xxicl{} fails.
\xxicl{} caused the model to reason using information that was weakly inferable yet incorrect.
}
\label{fig:output_neg_1}
\end{figure*}
\begin{figure*}[t]
\centering
\begin{tcolorbox}[title=Failure case of \xxicl{}, coltitle=black, colback=myred!30, colframe=myred, boxrule=0.5mm, arc=0.8mm, width=1.0\textwidth]
\small
\textbf{Premise:} Gyula Trebitsch (3 November 1914 - 12 December 2005) was a German film producer born in Budapest, Hungary. He was nominated in 1956 for the Academy Award for Best Foreign Language Film along with Walter Koppel for their film "The Captain of Kopenick".\\
\textbf{Hypothesis:} Gyula Trebitsch produced films in english.\\
\textbf{Label:} neutral\\

\textbf{Output of X-ICL}\\
Reason: The premise does not provide information about the language of the films produced by Gyula Trebitsch. It only mentions that he was nominated for an Academy Award for a film, but does not specify the language of that film or any other films he produced.\\
Label: neutral\\

\textbf{Output of \xxicl{}}\\
Possible Reasoning for entailment: The premise mentions that Gyula Trebitsch was nominated for an Academy Award for Best Foreign Language Film, which implies that he produced films in a language other than English, not in English.\\
Possible Reasoning for neutral: The premise provides information about Gyula Trebitsch's nomination for a foreign language film, but it does not specify whether he produced films in English or not, leaving the possibility open.\\
Possible Reasoning for contradiction: The premise indicates that Gyula Trebitsch was nominated for a Best Foreign Language Film, suggesting that his work was not in English, which contradicts the hypothesis.\\
Label: contradiction
\end{tcolorbox}
\caption{An example in which X-ICL provides the correct answer, whereas \xxicl{} fails.
\xxicl{} caused the model to reason using information that was inferable yet incorrect with respect to the gold label assigned.
}
\label{fig:output_neg_2}
\end{figure*}
\begin{figure*}[t]
\centering
\begin{tcolorbox}[title=Failure case of \xxicl{}, coltitle=black, colback=myred!30, colframe=myred, boxrule=0.5mm, arc=0.8mm, width=1.0\textwidth]
\small
\textbf{Premise:} Seven Seas to Calais (Italian: Il dominatore dei sette mari) is a 1962 Italian adventure film by Eastmancolor in CinemaScope, directed by Rudolph Maté (his final film) and Primo Zeglio. It stars Rod Taylor, Keith Michell and Edy Vessel. The film depicts the career of Sir Francis Drake.\\
\textbf{Hypothesis:} Rudolph Maté began directing films in 1960.\\
\textbf{Label:} neutral\\

\textbf{Output of X-ICL}\\
Reason: The premise states that "Seven Seas to Calais" was directed by Rudolph Maté in 1962, but it does not provide information about when he began directing films. Therefore, the hypothesis cannot be confirmed or denied based on the given information.\\
Label: neutral\\

\textbf{Output of \xxicl{}}\\
Possible Reasoning for entailment: The premise states that "Seven Seas to Calais" was directed by Rudolph Maté and released in 1962, which could imply that he began directing films around 1960.\\
Possible Reasoning for neutral: The premise provides information about the film "Seven Seas to Calais" and its directors but does not provide any information about when Rudolph Maté began his directing career.\\
Possible Reasoning for contradiction: The premise does not provide any information about when Rudolph Maté began directing films, so it cannot be concluded that he started in 1960. Additionally, Rudolph Maté had a long career in film before 1960, which contradicts the hypothesis.\\
Label: contradiction
\end{tcolorbox}
\caption{An example in which X-ICL provides the correct answer, whereas \xxicl{} fails.
\xxicl{} caused the model to rely on encyclopedic knowledge beyond the given context, which is not appropriate for the NLI task.
}
\label{fig:output_neg_3}
\end{figure*}
\begin{figure*}[t]
\centering
\begin{tcolorbox}[title=Failure case of \xxicl{}, coltitle=black, colback=myred!30, colframe=myred, boxrule=0.5mm, arc=0.8mm, width=1.0\textwidth]
\small
\textbf{Premise:} Oksa is a village in Jędrzejów County, Świętokrzyskie Voivodeship, in south-central Poland. It is the seat of the gmina (administrative district) called Gmina Oksa. It lies approximately 18 km north-west of Jędrzejów and 41 km south-west of the regional capital Kielce. The village has an approximate population of 890.\\
\textbf{Hypothesis:} The population of Oksa is large.\\
\textbf{Label:} contradiction\\

\textbf{Output of X-ICL}\\
Reason: The premise states that the population of Oksa is approximately 890, which is not considered large for a village. The term "large" is subjective, but typically a population of 890 would be considered small or average for a village.\\
Label: contradiction\\

\textbf{Output of \xxicl{}}\\
Possible Reasoning for entailment: The premise states that Oksa has an approximate population of 890, which could be considered large for a village.\\
Possible Reasoning for neutral: The premise provides the population of Oksa, but whether 890 is considered large is subjective and depends on the context or comparison to other villages.\\
Possible Reasoning for contradiction: The premise states that Oksa has an approximate population of 890, which is generally not considered large, especially when compared to cities or larger towns.\\
Label: neutral
\end{tcolorbox}
\caption{An example in which X-ICL provides the correct answer, whereas \xxicl{} fails.
In contrast to the other examples, \xxicl{} failed to utilize the inferable information in this case, which led the model to respond based on the absence of explicit information in the given context.
}
\label{fig:output_neg_4}
\end{figure*}

\end{document}